\documentclass[letterpaper]{article} 
\usepackage{aaai25}  
\usepackage{times}  
\usepackage{helvet}  
\usepackage{courier}  
\usepackage[hyphens]{url}  
\usepackage{graphicx} 
\usepackage{natbib}  
\usepackage{caption} 
\usepackage{algorithm}
\usepackage{xcolor}
\usepackage{amsmath}
\usepackage{multirow}
\usepackage{subfigure}
\usepackage{tikz}
\usepackage{amsfonts}
\usepackage{booktabs}
\usepackage{amssymb}
\usepackage[toc,title,page]{appendix}

\usepackage{algorithm}
\usepackage[noend]{algpseudocode}
\usepackage{amsfonts}
\usepackage{newfloat}
\usepackage{listings}
\newcommand{\update}[1]{#1}
\DeclareCaptionStyle{ruled}{labelfont=normalfont,labelsep=colon,strut=off} 
\floatstyle{ruled}
\newfloat{listing}{tb}{lst}{}
\floatname{listing}{Listing}
\title{CSformer: Combining Channel Independence and Mixing for Robust Multivariate Time Series Forecasting}
\author{
    Haoxin Wang\textsuperscript{\rm 1}, Yipeng Mo\textsuperscript{\rm 1}, Kunlan Xiang\textsuperscript{\rm 2}, Nan Yin\textsuperscript{\rm 1}, Honghe Dai\textsuperscript{\rm 1}\\
    Bixiong Li\textsuperscript{\rm 3}, Songhai Fan\textsuperscript{\rm 4}, Site Mo\textsuperscript{\rm 1}\thanks{Corresponding author.}
}
\affiliations{
    \textsuperscript{\rm 1}College of Electrical Engineering, Sichuan University
    \textsuperscript{\rm 2}University of Electronic Science and Technology of China\\
    \textsuperscript{\rm 3}College of Architecture and Environment, Sichuan University
    \textsuperscript{\rm 4}State Grid Sichuan Electric Power Research Institute\\

    \{whx1122, moyipeng\}@stu.scu.edu.cn, mosite@126.com
}

\usepackage{bibentry}

\begin{document}

\maketitle

\begin{abstract}
In the domain of multivariate time series analysis, the concept of channel independence has been increasingly adopted, demonstrating excellent performance due to its ability to eliminate noise and the influence of irrelevant variables. However, such a concept often simplifies the complex interactions among channels, potentially leading to information loss. To address this challenge, we propose a strategy of channel independence followed by mixing. Based on this strategy, we introduce \textbf{CSformer}, a novel framework featuring a two-stage multiheaded self-attention mechanism. This mechanism is designed to extract and integrate both \textbf{C}hannel-specific and \textbf{S}equence-specific information. Distinctively, CSformer employs parameter sharing to enhance the cooperative effects between these two types of information. Moreover, our framework effectively incorporates sequence and channel adapters, significantly improving the model's ability to identify important information across various dimensions. Extensive experiments on several real-world datasets demonstrate that CSformer achieves state-of-the-art results in terms of overall performance.
\end{abstract}

%

\section{Introduction}

Time series forecasting is vital in areas such as traffic management \citep{cirstea2022towards,yin2016forecasting,qin2023spatio}, power systems \citep{stefenon2023wavelet,wang2023forecasting,mo2024powerformer}, and healthcare \citep{ahmed2023multivariate,alshanbari2023implementation,sen2022forecasting}. However, it faces significant challenges due to the complex long-term dependencies and variable interrelations inherent in time series data. These difficulties have propelled multivariate time series forecasting (MTSF) to the forefront of research in these domains.

 Recently, many deep learning models have been applied to MTSF, especially Transformer-based models, such as Informer ~\citep{Informer}, Autoformer ~\citep{Autoformer}, and Fedformer ~\citep{fedformer}. These models have demonstrated improved predictive performance by refining the attention mechanism. However, recent research has raised questions regarding the suitability of Transformer models for MTSF task, proposing a straightforward linear model, DLinear ~\citep{DLinear}, that surpasses traditional Transformer models. Given the Transformer's success in other domains (NLP, CV, etc.) ~\citep{devlin2018bert,brown2020language,dosovitskiy2020image}, researchers have shifted their focus from modifying the attention structure within Transformers to altering the input data. iTransformer ~\citep{liu2023itransformer} treats sequences from each channel as a single token, embedding them along the sequence dimension and applying attention across the channel dimension. PatchTST ~\citep{PatchTST} uses a channel independent approach. It divides the input data into patches before feeding it into the Transformer and then embedding each patch as a token. These approaches have reinstated Transformer methods to a prominent position.
\begin{figure}[t]
\begin{center}
\includegraphics[width=1\columnwidth]{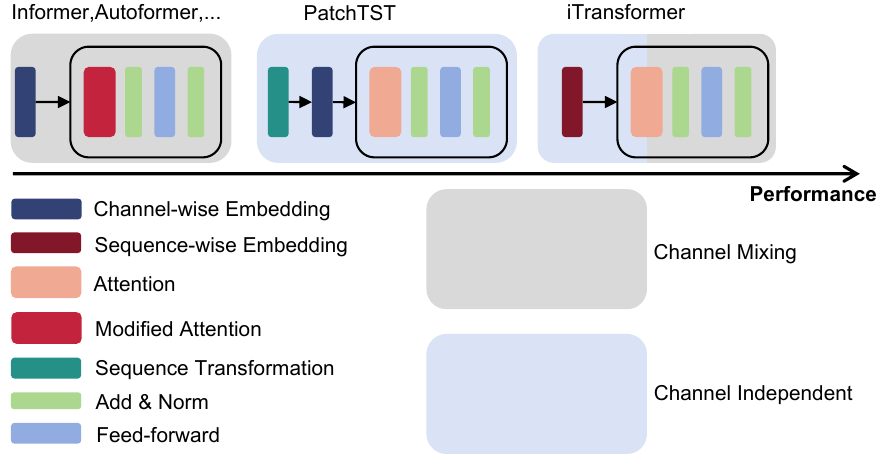}
 \caption{Transformer-based models categorized by channel independence and channel mixing.}
\label{fig:channel}
\end{center}
\end{figure}
We summarize the Transformer-based model based on the channel processing approach. As shown in Figure~\ref{fig:channel}, traditional models like Informer and Autoformer \citep{Informer, Autoformer} directly embed variables, leading to channel mixing and potential noise introduction, especially from varied sensors and distributions. To eliminate these defects, PatchTST \citep{PatchTST} adopts a channel-independent approach, risking information loss due to reduced physical intuitiveness. These observations suggest that neither direct channel independence nor mixing is optimal. To address this, hybrid methods like iTransformer \citep{liu2023itransformer} combine sequence embedding with channel-wise attention, offering a balanced approach. However, embedding methods of iTransformer may lead to erasure of inter-sequence correlations and treat the extraction of sequence information and channel information as separate non-interactive processes.
\begin{figure}[ht]

{
\includegraphics[width=0.42\textwidth]{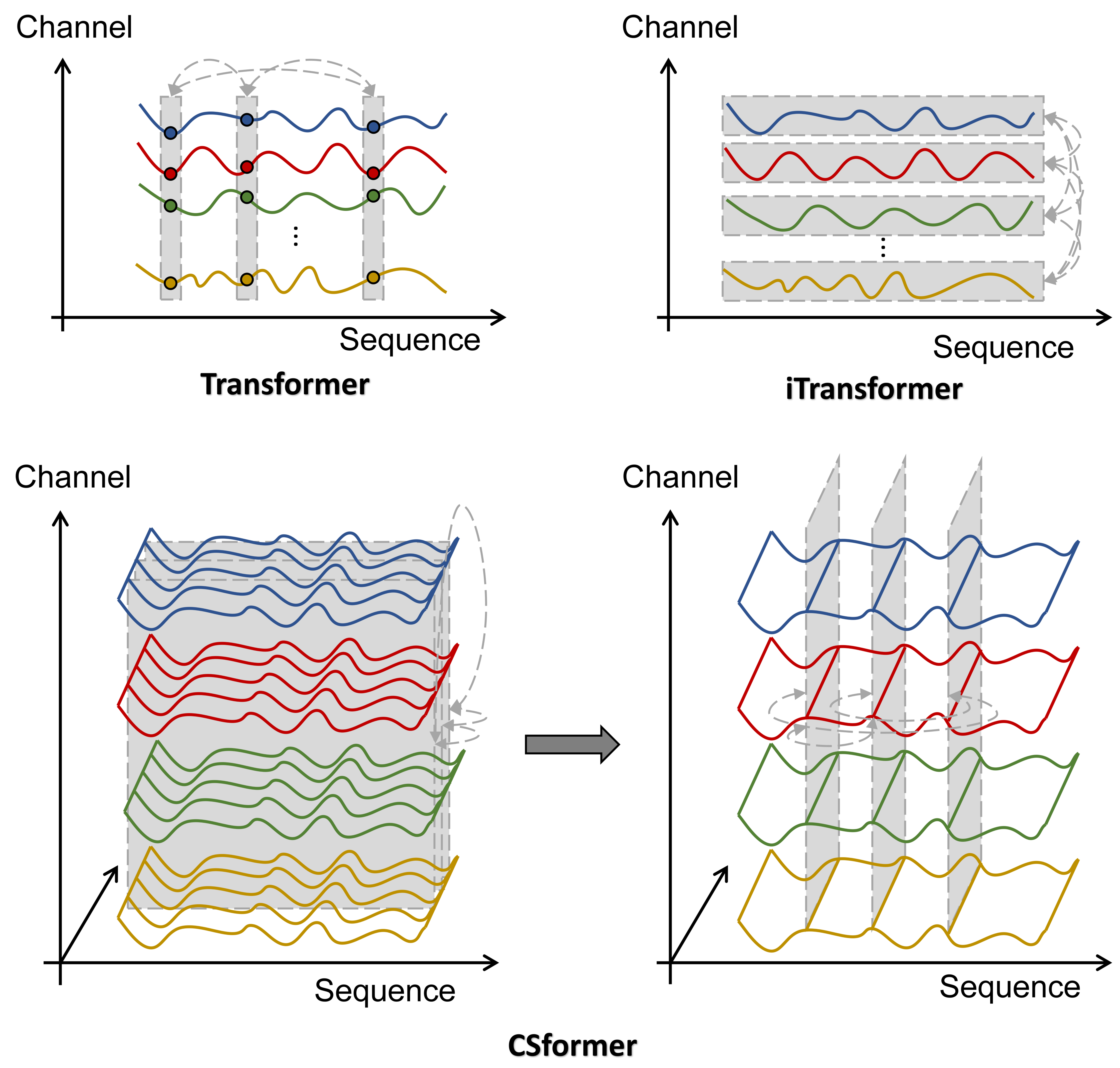}
\caption{Structural Comparison of Transformer (top left), iTransformer (top right), and proposed CSformer (bottom): In the illustration, we compare the architectures of Transformer (top left) and iTransformer (top right) with the proposed CSformer (bottom). While Transformer and iTransformer employ attention mechanisms separately in the sequence and channel dimensions, CSformer diverges by embedding sequences into a high-dimensional space. Consequently, CSformer performs attention independently in both channel and sequence dimensions.}\label{fig:motivation}
}
\end{figure}

Drawing on the aforementioned motivation, we believe an effective combination of channel independence and channel mixing is crucial for mining more robust features in multivariate time-series data. In this paper, We propose CSformer, a model adept at extracting and combining sequence and channel information without modifying Transformer's attention mechanism. As depicted in Figure \ref{fig:motivation}, our approach introduces a dimensionality-augmented embedding technique that enhances the sequence dimensionality while preserving the original data's integrity. We then use a unified multiheaded self-attention (MSA) mechanism to apply different attentions on sequence and channel dimensions respectively, employing a shared-parameter approach to realize the interplay of information from different dimensions. After each MSA mechanism, we include an adapter to ensure that different features are extracted by the two-stage self-attention mechanism. The primary contributions of our work are outlined as follows:

\begin{itemize}
  \item To enhance the extraction capabilities of channel and sequence information, we propose \textit{CSformer}, a two-stage attention Transformer model. This model efficiently captures the sequential and channel information of multivariate time series while minimizing the increase in model parameters, enabling effective information interaction.
  \item CSformer achieves state-of-the-art (SOTA) performance on diverse real-world datasets, covering domains such as electricity and weather. Extensive ablation studies further validate the model's design rationality.
  \item We propose a new training strategy for MTSF: channel independence followed by channel mixing, providing a new insight for future research.

\end{itemize}
\begin{figure*}[htbp]
\centering
{
\includegraphics[width=0.92\textwidth]{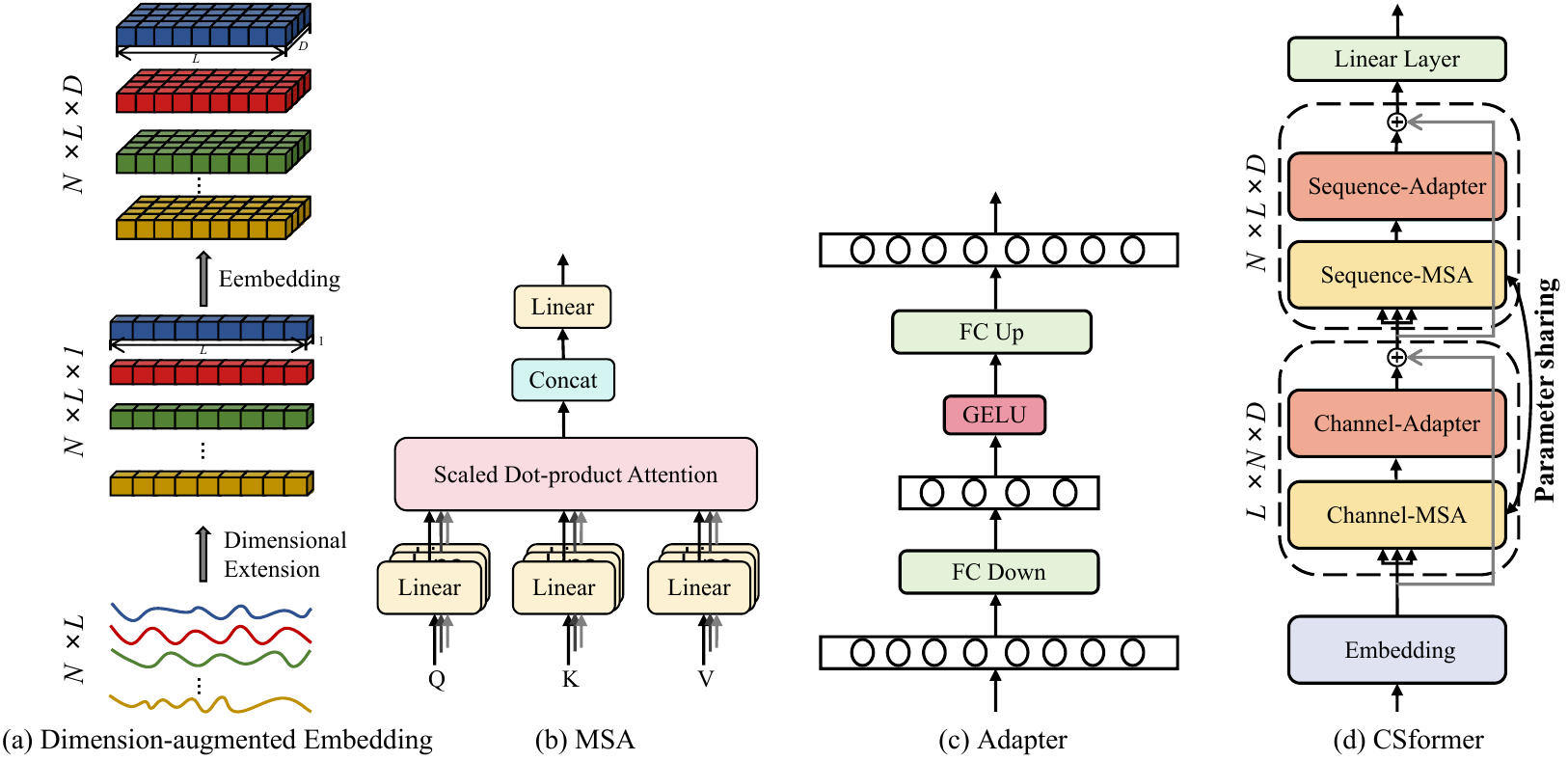}
\caption{We present the overall framework of CSformer (d). Initially, the input sequence undergoes a dimensional expansion operation before embedding (a). This dimensional transformation allows the standard MSA (b) to be adapted separately for channels and sequences (c). Note that Channel-MSA and Sequence-MSA share weights but are applied to different input dimensions.}\label{fig:framework}
}
\end{figure*}

\section{Related Work}

Time series forecasting models are broadly classified into statistical and deep learning models. Statistical models, such as ARIMA \citep{bartholomew1971time} and Exponential Smoothing \citep{hyndman2008forecasting}, are noted for their simplicity and efficacy in capturing temporal dynamics, but mainly focus on univariate forecasting, neglecting additional covariates. Conversely, deep learning models, including RNN-based \citep{medsker2001recurrent,graveslong,dey2017gate,mo2023timesql}, CNN-based \citep{bai2018empirical,wang2022micn, Timesnet}, and Transformer-based \citep{Informer, Autoformer,fedformer}, have been prominent in MTSF for their comprehensive capabilities. They aim to improve prediction accuracy by integrating temporal and inter-channel information. Recent studies, however, indicate focusing solely on sequence information may be more effective. Therefore, deep learning models in MTSF can be categorized as channel-independent or channel-mixing based on their handling of inter-channel information.

\paragraph{Channel-independent Models} The channel-independent models process each time series channel (variable) independently, disregarding any interdependencies or interactions among different channels. Intuitively, this simplistic approach may not yield optimal results in certain contexts due to its failure to consider the potentially complex interrelationships within time series data. Despite this, significant 
performance enhancements have been observed with such models. DLinear \citep{DLinear}, a proponent of channel-independent architectures, employs a singular linear model along the sequence dimension and surpasses the performance of all contemporary state-of-the-art Transformer-based models. PatchTST \citep{PatchTST}, another exemplar of channel-independent models, initially divides a single time series into segments, treats each segment as a token, and then processes these segments through a transformer model to extract temporal information.

\paragraph{Channel-mixing Models} The channel-mixing models are characterized by approaches in which the analysis of one time series channel is influenced by other channels. These models integrate interdependencies among multiple time series variables, facilitating richer and more accurate representations of data patterns and trends. Previous research \citep{li2019enhancing,Informer,Autoformer,fedformer} mainly concentrated on reducing the computational complexity of the attention mechanism but addressed inter-channel relationships only through basic embedding operations. This approach may not fully capture mutual information, and at times, it might inadvertently introduce noise. To address this limitation, Crossformer \citep{Crossformer} introduces a two-stage attention layer that leverages both cross-dimensional and cross-temporal dependencies. Similarly, TSMixer \citep{chen2023tsmixer} applies linear layers alternately across time and channel dimensions, thus enabling the extraction of temporal and cross-variate information. In a more recent development, iTransformer \citep{liu2023itransformer} conducts embedding operations along the time dimension and then utilizes the attention mechanism to extract information between variables, leading to notable performance enhancements.

As highlighted in PatchTST \citep{PatchTST}, improperly handling cross-dependency extraction can lead to the introduction of noise. Our objective with CSformer is to devise a method that effectively utilizes cross-variable information while concurrently adequately extracting temporal data. Additionally, considering that advancements in the attention mechanism have not substantially enhanced efficiency in practical applications \citep{DLinear}, this paper maintains adherence to the vanilla attention mechanism architecture.

\section{CSformer}
In this section, we present our novel model, the CSformer, which is capable of concurrently learning channel and sequence information. Figure~\ref{fig:framework} illustrates the overall architecture of CSformer. We first enhance multivariate time series data using dimension-augmented embedding. Then, we apply a two-stage MSA mechanism to extract channel and sequence features, sharing parameters to facilitate interaction between dimensions. Finally, we introduce channel/ sequence adapters for each output to strengthen their capability to learn information across different dimensions.

\subsection{Preliminaries}
In the context of multivariate time series forecasting, the historical sequence of inputs is denoted as $\mathbf{X}=\{\mathbf{x}_1,\ldots,\mathbf{x}_L\}\in\mathbb{R}^{N\times L}$, where $N$ represents the number of variables (channels) and $L$ represents the length of the historical sequence. Let $\mathbf{x}_i \in\mathbb{R}^{N}$ represent the values of the various variables at the $i$-th time point. Let $\mathbf{X}^{(k)} \in\mathbb{R}^{L}$ denote the sequence of the $k$-th variable. Assuming our prediction horizon is $T$, the predicted result is denoted as $\mathbf{\Hat{X}}=\{\mathbf{x}_{L+1},\ldots,\mathbf{x}_{L+T}\}\in\mathbb{R}^{N\times T}$. Considering a model $\mathbf{f_{\theta}}$, where $\theta$ represents the model parameters, the overall process of multivariate time series forecasting can be abstracted as $\mathbf{f_{\theta}(X)}\to \mathbf{\hat{X} }$.

\subsection{Reversible Instance Normalization}

In practical scenarios, time-series data distributions often shift due to external factors, leading to data distribution bias. To mitigate this, Reversible Instance Normalization (ReVIN) \citep{kim2021reversible} has been introduced. As delineated in Equation ~\ref{equ:revin}, ReVIN normalizes each input variable's sequence, mapping it to an alternative distribution. This process artificially minimizes distributional disparities among univariate variables in the original data. Subsequently, the original mean, variance, and learned parameters are reinstated in the output predictions.

\begin{equation}\scriptsize
\begin{aligned}
\operatorname{ReVIN}(\mathbf{X}) = \Bigg\{ & \mathbf{\gamma }_{k}\frac{\mathbf{X}^{(k)} - \operatorname{Mean}(\mathbf{X}^{(k)})}{ \sqrt{\operatorname{Var}(\mathbf{X}^{(k)})+\varepsilon }}+
& \mathbf{\beta }_{k} \bigg| \ k=1,\cdots,N \Bigg\},
\label{equ:revin}
\end{aligned}
\end{equation}
where $\varepsilon$ refers to an infinitesimal quantity that prevents the denominator from being 0. $\mathbf{\beta}\in\mathbb{R}^{N}$ is the parameter for performing the normalization and $\gamma\in\mathbb{R}^{N}$ is the learnable parameter for performing the affine transformation.

\subsection{Dimension-augmented Embedding} PatchTST \citep{PatchTST} segments input sequences into temporal blocks, treating each as a Transformer token. TimesNet \citep{Timesnet} uses a fast Fourier transform (FFT) to discern sequence cycles, reshaping them into 2D vectors for CNN processing. While these methods increase sequence dimensionality, they risk information loss (see Appendix B). To address this issue, we introduce a direct sequence embedding approach, which is inspired by word embedding techniques \citep{word2vec,yi2023frequency}. To maintain input data integrity, we first apply a dimension-augmentation operation: $\mathbf{X}\in\mathbb{R}^{N\times L}\to\mathbf{X}\in\mathbb{R}^{N\times L\times 1}$. Then, we multiply the augmented sequence element-wise with a learnable vector $ \mathbf{\nu}\in\mathbb{R}^{1\times D}$, producing the embedded output $\mathbf{H}\in\mathbb{R}^{N\times L\times D} = \mathbf{X} \times \mathbf{\nu}$. This method facilitates dimensionality enhancement and embedding without distorting the original input's intrinsic information, setting the stage for the subsequent two-stage MSA detailed later.

\subsection{Two-stage MSA}

The CSformer consists of $M$ blocks. 
Due to the dynamic weighting characteristics of the self-attention mechanism, each block employs a two-stage MSA using shared parameters. A brief exposition of this characteristic is as follows: with input data $\mathbf{H}\in\mathbb{R}^{S\times D}$, where $S$ represents $N$ or $L$, we define $\mathbf{Q},\mathbf{K} \in\mathbb{R}^{S\times D_{k}}$ as query and key. The attention score $\mathbf{A}\in \mathbb{R}^{S \times S}$ can be derived as $\operatorname{Softmax}(\mathbf{Q}\mathbf{K}^\top / \sqrt{D_k})$, where $\mathbf{Q}=\mathbf{H}\mathbf{W_{Q}}$ and $\mathbf{K}=\mathbf{H}\mathbf{W_{K}}$, with $\mathbf{W_{Q}},\mathbf{W_{K}}\in \mathbb{R}^{D \times D_{k}}$ as projection matrices of query and key. By the associative law of matrix multiplication, $\mathbf{A}$ can be written as $\operatorname{Softmax}(\mathbf{H}\mathbf{W}\mathbf{H}^\top / \sqrt{D_k})$, where $\mathbf{W}=\mathbf{W_{Q}}\mathbf{W_{K}}^\top$. Since $\mathbf{W},D_{k}$ are fixed values, $\mathbf{A}$ dynamically adapts to changes in input data $\mathbf{H}$. This characteristic implies that with varying dimensional inputs, the attention scores adaptively adjust, even under a shared parameter method.
Following the application of MSA, an adapter is incorporated to optimize the discriminative learning of both channel and sequence information. Detailed discussions of the MSA and adapter components are outlined in subsequent sections.

\paragraph{Channel MSA} The channel MSA in the first stage is similar to the iTransformer ~\citep{liu2023itransformer}, both employing MSA along the channel dimension. However, a distinction arises in the treatment of the time series, while the iTransformer considers the entire time series as a single token, herein, we apply channel-wise attention at each time step to discern inter-channel dependencies. Let $\mathbf{H}_c\in \mathbb{R}^{L\times N\times D}$
denote the input subjected to the channel-wise MSA. The output post-attention $\mathbf{Z}_c \in \mathbb{R}^{L\times N\times D}$ is formulated as:

\begin{equation}\label{equ:norm}
    \mathbf{Z}_c = \operatorname{MSA}(\mathbf{H}_c).
\end{equation}

The shared parameters between channel and sequence MSAs in CSformer pose challenges in learning across two dimensions simultaneously. Inspired by fine-tuning technique in NLP's large-scale model adaptation ~\citep{houlsby2019parameter,hu2022lora}, we integrate adapter technology ~\citep{houlsby2019parameter}. This comprises two fully connected layers interspersed with an activation function. The first layer downscales input features, followed by activation, and the second layer reverts the representation to its original dimensionality. This iterative process enables the model to capture channel representations more effectively. Thus, after channel information extraction, the model output is delineated as follows:

\begin{equation}\label{equ:norm}
    \mathbf{A}_c = \operatorname{Adapter}(\operatorname{Norm}(\mathbf{Z}_c))+\mathbf{H}_c,
\end{equation}
where $\mathbf{A}_c\in \mathbb{R}^{L\times N \times D}$ denotes the output after channel information extraction, with $\operatorname{Norm}$ indicating the normalized MSA output via batch normalization. To mitigate gradient challenges, the adapter's output is additively fused with $\mathbf{H}_c$, enhancing overall framework stability.

\paragraph{Sequence MSA}
Transformer ~\citep{Transformer} and iTransformer ~\citep{liu2023itransformer} models traditionally concentrate on sequence or channel dimensions. Crossformer ~\citep{Crossformer} introduces a novel approach, applying attention first to the sequence, then the channel dimension. However, its independent two-stage mechanism may result in each MSA only being able to focus on specific aspects of information, reducing the ability to fuse information.

\begin{table*}[htbp]
  
  \centering
\renewcommand{\arraystretch}{1.2}
	\resizebox{\linewidth}{!}{
  \begin{tabular}{c|cc|cc|cc|cc|cc|cc|cc|cc|cc|cc|cc|cc}
    \toprule
    
    \multicolumn{1}{c}{\multirow{1}{*}{Models}} & 
    \multicolumn{2}{c}{\rotatebox{0}{{\textbf{CSformer}}}} &
    \multicolumn{2}{c}{\rotatebox{0}{{iTransformer}}} &
    \multicolumn{2}{c}{\rotatebox{0}{{PatchTST}}} &
    \multicolumn{2}{c}{\rotatebox{0}{{Crossformer}}} &
    \multicolumn{2}{c}{\rotatebox{0}{{SCINet}}} &
    \multicolumn{2}{c}{\rotatebox{0}{{TiDE}}} &
    \multicolumn{2}{c}{\rotatebox{0}{{{TimesNet}}}} &
    \multicolumn{2}{c}{\rotatebox{0}{{DLinear}}} &
    \multicolumn{2}{c}{\rotatebox{0}{{FEDformer}}} &
    \multicolumn{2}{c}{\rotatebox{0}{{Stationary}}} &
    \multicolumn{2}{c}{\rotatebox{0}{{Autoformer}}} &
    \multicolumn{2}{c}{\rotatebox{0}{{Informer}}} \\
    
    \cmidrule(lr){2-3} \cmidrule(lr){4-5}\cmidrule(lr){6-7} \cmidrule(lr){8-9}\cmidrule(lr){10-11}\cmidrule(lr){12-13} \cmidrule(lr){14-15} \cmidrule(lr){16-17} \cmidrule(lr){18-19} \cmidrule(lr){20-21} \cmidrule(lr){22-23} \cmidrule(lr){24-25}
    \multicolumn{1}{c}{Metric}  & {MSE} & {MAE}  & {MSE} & {MAE}  & {MSE} & {MAE}  & {MSE} & {MAE}  & {MSE} & {MAE}  & {MSE} & {MAE} & {MSE} & {MAE} & {MSE} & {MAE} & {MSE} & {MAE} & {MSE} & {MAE} & {MSE} & {MAE} & {MSE} & {MAE}\\
    \toprule
    \multirow{1}{*}{\update{{{ETTm1}}}}
    & \textbf{{0.385}} & \textbf{{0.400}} & {0.407} & {0.410}  & \underline{{0.387}} & \textbf{{0.400}} & {0.513} & {0.496} & {0.419} & {0.419} &{0.400} &\underline{{0.406}} &{{0.403}} &{{0.407}} & {0.485} & {0.481}  &{0.448} &{0.452} &{0.481} &{0.456} &{0.588} &{0.517} &{0.961} &{0.734} \\
    \midrule
    \multirow{1}{*}{\update{{{ETTm2}}}}
    & \underline{{0.282}} &\underline{{0.331}}& {{0.288}} & {{0.332}}  & \textbf{{0.281}} & \textbf{{0.326}} & {0.757} & {0.610} & {0.358} & {0.404} &{{0.291}} &{{0.333}} &{0.350} &{0.401} & {0.571} & {0.537} &{0.305} &{0.349} &{0.306} &{0.347} &{0.327} &{0.371} &{1.410} &{0.810} \\
    \midrule
    \multirow{1}{*}{{\update{{ETTh1}}}}
    & \textbf{{{0.429}}} & \textbf{{{0.432}}}& {{0.454}} & \underline{{0.447}} & {0.469} & {0.454} & {0.529} & {0.522} & {0.541} & {0.507} &{0.458} &{{0.450}} &{{0.456}} &{{0.452}} & {0.747} & {0.647} &\underline{{0.440}} &{0.460} &{0.570} &{0.537} &{0.496} &{0.487}  &{1.040} &{0.795} \\
    \midrule
    \multirow{1}{*}{{{ETTh2}}}  
    & {\textbf{0.364}} & {\textbf{0.392}}
    & \underline{{0.383}} & \underline{{0.407}} & {0.387} & \underline{{0.407}} & {0.942} & {0.684} & {0.954} & {0.723} & {0.611} & {0.550}  &{{0.414}} &{{0.427}} &{0.559} &{0.515} &{{0.437}} &{{0.449}} &{0.526} &{0.516} &{0.450} &{0.459} &{4.431} &{1.729} \\
    \midrule
    \multirow{1}{*}{{{Electricity}}} 
    & {\textbf{0.176}} & {\textbf{0.270}} & \underline{{0.178}} & \textbf{{0.270}} & {0.216} & {0.304} & {0.244} & {0.334} & {0.268} & {0.365} & {0.251} & {0.344} &{{0.192}} &\underline{{{0.295}}} &{0.212} &{0.300} &{0.214} &{0.327} &{{0.193}} &{{0.296}} &{0.227} &{0.338} &{0.311} &{0.397} \\
    \midrule
    \multirow{1}{*}{{{Solar Energy}}} 
    &{\textbf{0.230}} &\underline{{0.270}}&\underline{{0.233}} &{\textbf{0.262}} &{0.270} &{0.307} &{0.641} &{0.639} &{0.282} &{0.375} &{0.347} &{0.417} &{0.301} &{0.319} &{0.330} &{0.401} &{0.291} &{0.381} &{0.261} &{0.381} &{0.885} &{0.711} &{0.235} &{0.280}\\
    \midrule
    \multirow{1}{*}{{{Weather}}} 
    & {\textbf{0.250}} & {\underline{0.280}} & {\underline{0.258}} & {\textbf{0.279}} & {0.259} & {0.281} & {0.259} & {0.315} & {0.292} & {0.363} & {0.271} & {0.320} &{{0.259}} &{{0.287}} &{0.265} &{0.317} &{0.309} &{0.360} &{0.288} &{0.314} &{0.338} &{0.382} &{0.634} &{0.548} \\
    \midrule
     \multicolumn{1}{c|}{{{$1^{\text{st}}$ Count}}} & {\textbf{6}} & {\textbf{4}} & {0} & {\underline{3}} & {\underline{1}} & {2} & {0} & {0} & {0} & {0} & {0} & {0} & {0} & {0} & {0} & {0} & {0} & {0} & {0} & {0} & {0} & {0} & {0} & {0}\\
    \bottomrule
  \end{tabular}
  }
  \caption{Multivariate forecasting results. We compare extensive competitive models under different prediction lengths following the setting of iTransformer. The input sequence length is set to 96 for all baselines. Results are averaged from all prediction lengths. The best results are in \textbf{bold} and the second best are \underline{underlined}. Full results are listed in Appendix D.}

  \label{tab:results}
\end{table*}

To address this challenge, we present a novel method: the reuse of parameters derived from MSA applied along the channel dimension for sequence modeling. Specifically, the output $\mathbf{A}_c\in \mathbb{R}^{L\times N \times D}$
from the channel MSA undergoes a reshape operation to seamlessly transition into the input $\mathbf{H}_s\in \mathbb{R}^{N\times L \times D}$
for the subsequent sequence MSA. This is designed to establish the interactions between channel and sequence representations and to extract the implicit associations between channels and sequences.

When the input $\mathbf{H}_s$ is fed into the sequence MSA, the output value $\mathbf{Z}_s$ can be formulated as:

\begin{equation}\label{equ:norm}
    \mathbf{Z}_s = \operatorname{MSA}(\mathbf{H}_s).
\end{equation}
The MSA in this study is a shared layer, distinctively applied across different input dimensions. This efficient operation substantially enhances both sequence and channel modeling capabilities while maintaining the number of parameters.

In line with the channel MSA, a subsequent adapter is introduced post-reused MSA layer for sequence feature accommodation. The architecture of this sequence adapter is analogous to that of the channel adapter, described as follows:

\begin{equation}\label{equ:norm}
    \mathbf{A}_s = \operatorname{Adapter}(\operatorname{Norm}(\mathbf{Z}_s))+\mathbf{H}_s,
\end{equation}
where $\mathbf{A}_s \in \mathbb{R}^{N\times L\times D}$ denotes the output values of the sequence adapter.
The integration of this approach is pivotal for imparting distinctiveness to sequence and channel features. The model can attain refined differentiation by utilizing a lightweight adapter, bolstering functionality without substantially increasing complexity.

\subsection{Prediction}
After two-stage MSA, we reduce the data's complexity by squeezing the sequence into a lower dimension, forming $\mathbf{Z} \in \mathbb{R}^{N\times (L*D)}$. This step smartly packs the features, keeping the important information. Next, we use a linear layer for the final prediction, getting prediction result $\mathbf{\Hat{X}} \in \mathbb{R}^{N\times T}$. 

\section{Experiments}

In this section, a wide range of experiments was meticulously conducted to evaluate the CSformer model. The experiments were designed with a detailed examination of the rationality of the model's architecture. This thorough evaluation aimed not only to examine the model's performance across various dimensions but also to gain deeper insights into the principles behind its design.

\subsection{Datasets} 
To evaluate the performance of CSformer, we employed commonly used datasets for various domains in the multivariate long-term forecasting tasks, including ETT (ETTh1, ETTh2, ETTm1, ETTm2), Weather, Electricity, and Solar Energy. 





\subsection{Baseline}
We have selected a multitude of advanced deep learning models that have achieved SOTA performance in multivariate time series forecasting. Our choice of Transformer-based models includes the Informer ~\citep{Informer}, Autoformer ~\citep{Autoformer}, Stationary ~\citep{Stationary}, FEDformer ~\citep{fedformer}, Crossformer ~\citep{Crossformer}, PatchTST ~\citep{PatchTST}, and the leading model, iTransformer ~\citep{liu2023itransformer}. Notably, Crossformer ~\citep{Crossformer} shares conceptual similarities with our proposed CSformer. Furthermore, recent advancements in MLP-based models have yielded promising predictive outcomes. In this context, we have specifically selected representative models such as DLinear ~\citep{DLinear} and TiDE ~\citep{Tide} for our comparative evaluation. Additionally, for a more comprehensive comparison, we also selected TCN-based models, which include SCINet ~\citep{SCINet} and TimesNet ~\citep{Timesnet}. 

\subsection{Experimental Setting}
We adhered to the experimental configurations employed in iTransformer ~\citep{liu2023itransformer}. For all datasets, the look-back window length $L$ is set to 96, and the prediction length is designated as $T \in \left \{ 96,192,336,720 \right \} $. All experiments were conducted on an NVIDIA A40 GPU, utilizing the PyTorch ~\citep{Pytorch} framework. The mean squared error (MSE) serves as the chosen loss function, with both MSE and mean absolute error (MAE) employed as evaluation metrics to gauge the efficacy of the prediction results.

\subsection{Main results} As shown in Table 
~\ref{tab:results}, our proposed CSformer achieves the overall best performance across all datasets. Specifically, it secured the top position in 6 out of all the MSE metrics and 4 out of the MAE metrics. This leading number of top-1 rankings across all models underlines the superiority and effectiveness of our approach. It is worth noting that, compared to the latest SOTA model, iTransformer ~\citep{liu2023itransformer}, we have further explored the potential of Transformer-based models, thereby achieving additional performance improvements. Besides, CSformer has outperformed Crossformer ~\citep{Crossformer}, a model with a similar design approach, by a large margin. We believe that this may be due to the fact that Crossformer does not interact when extracting cross-sequence and cross-channel information. In addition, our model has demonstrated significant improvements over the channel-independent PatchTST ~\citep{PatchTST}. This highlights the need to more effectively extract dependencies between variables.

\begin{figure}[h]
\begin{center}
\includegraphics[width=1\columnwidth]{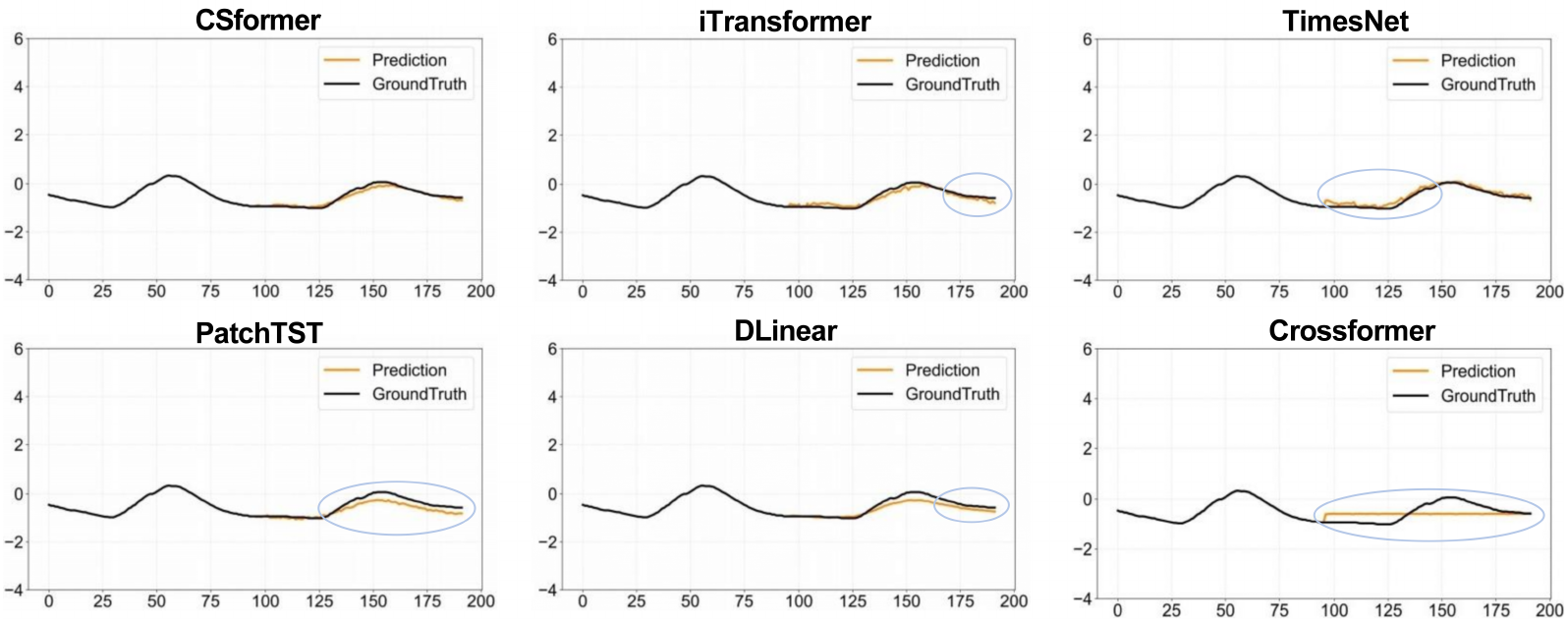}
\caption{Visualization of input-96-predict-96 results on the ETTm2 dataset.}
\label{fig:visualization}
\end{center}
\end{figure}

Figure ~\ref{fig:visualization} shows that CSformer outperforms other models in prediction accuracy, stability, detail capture, and trend tracking, likely due to its superior integration of features across sequences and channels.

\subsection{Ablation Study}
\paragraph{Two-stage MSA}
To evaluate the importance of each MSA component in the two-stage MSA framework, we removed one MSA (including its adapter) for comparative analysis. As evidenced by Table ~\ref{tab:MSA}, CSformer demonstrates superior overall performance. Notably, retaining only the sequence MSA also yields satisfactory results, attributable to the resulting channel independence.

\begin{table}[h]
	\centering
	\resizebox{1\linewidth}{!}{
		\begin{tabular}{c|cc|cc|cc}
    \toprule
    \multirow{1}{*}{{Datasets}} & 
    \multicolumn{2}{c}{\rotatebox{0}{{ETTh1}}} &
    \multicolumn{2}{c}{\rotatebox{0}{{Weather}}} &
    \multicolumn{2}{c}{\rotatebox{0}{{Electricity}}} \\
    \cmidrule(lr){2-3} \cmidrule(lr){4-5}\cmidrule(lr){6-7} 
    Metric & {MSE} & {MAE}  & {MSE} & {MAE}  & {MSE} & {MAE} \\
    \toprule
    {w/o Channel MSA} & 0.431 & 0.433 & \textbf{0.248} & 0.278 & 0.197 & 0.282\\
    {w/o Sequence MSA} & 0.443 & \textbf{0.432} & 0.259 & \textbf{0.272} & 0.200 & 0.286 \\
    \cmidrule(lr){1-7}
    {CSformer} & \textbf{0.429} &\textbf{0.432} & 0.250 & 0.280 & \textbf{0.176} & \textbf{0.270} \\
    \bottomrule
  \end{tabular}
	}
 \caption{Ablation of two-stage MSA on three real-world datasets with MSE and MAE metrics.}
	\label{tab:MSA}
\end{table}

Comparing parameter sharing between channel MSA and sequence MSA with their non-shared counterparts reveals that shared parameters generally yield better prediction outcomes (see Table ~\ref{tab:paremeter}). This improvement is attributed to the interaction between channel and sequence information. Although the performance of the non-shared method does not degrade significantly compared to the performance of the parameter-sharing method, it significantly increases the number of parameters in the model, which may lead to parameter redundancy. Therefore, parameter sharing is a better choice.

\begin{table}[h]
	\centering
	\resizebox{1\linewidth}{!}{
		\begin{tabular}{c|cc|cc|cc}
    \toprule
    \multirow{1}{*}{{Datasets}} & 
    \multicolumn{2}{c}{\rotatebox{0}{{ETTh1}}} &
    \multicolumn{2}{c}{\rotatebox{0}{{Weather}}} &
    \multicolumn{2}{c}{\rotatebox{0}{{Electricity}}} \\
    \cmidrule(lr){2-3} \cmidrule(lr){4-5}\cmidrule(lr){6-7} 
    Metric & {MSE} & {MAE}  & {MSE} & {MAE}  & {MSE} & {MAE} \\
    \toprule
    {w/o Parameter Sharing} & 0.440 & 0.438 & 0.256 & 0.282 & 0.189 & 0.280\\
    \cmidrule(lr){1-7}
    {Parameter Sharing} & \textbf{0.429} &\textbf{0.432} & \textbf{0.250} & \textbf{0.280} & \textbf{0.176} & \textbf{0.270} \\
    \bottomrule
  \end{tabular}
	}
 \caption{Ablation of parameter sharing on three real-world datasets with MSE and MAE metrics.}

	\label{tab:paremeter}
\end{table}

Finally, we thoroughly tested the influence of varying the order of application between sequence MSA and channel MSA on the predictive performance of our model.

As shown in Table ~\ref{tab:order}, the experimental results suggests a superior efficacy when channel MSA precedes sequence MSA (C \(\to\) S). This ordered approach facilitates the initial establishment of inter-channel interactions and connections, thereby providing a comprehensive base of information. Such information is invaluable in laying down a relevant understanding of channel dynamics. This initial comprehension of inter-channel relationships significantly enriches the subsequent analysis of temporal evolution of these variables in sequence MSA. This insight highlights the critical nature of the order of operations in MSA applications, especially in the context of enhancing predictive accuracy in complex datasets.

\begin{table}[htbp]
	\centering
	\resizebox{1\linewidth}{!}{
		\begin{tabular}{c|cc|cc|cc}
    \toprule
    \multirow{1}{*}{{Datasets}} & 
    \multicolumn{2}{c}{\rotatebox{0}{{ETTh1}}} &
    \multicolumn{2}{c}{\rotatebox{0}{{Weather}}} &
    \multicolumn{2}{c}{\rotatebox{0}{{Electricity}}} \\
    \cmidrule(lr){2-3} \cmidrule(lr){4-5}\cmidrule(lr){6-7} 
    Metric & {MSE} & {MAE}  & {MSE} & {MAE}  & {MSE} & {MAE} \\
    \toprule
    {S \(\to\) C} & 0.443 & 0.442 & 0.250 & 0.280 & 0.178 & 0.274\\
    \cmidrule(lr){1-7}
    {C \(\to\) S} & \textbf{0.429} &\textbf{0.432} & \textbf{0.250} & \textbf{0.280} & \textbf{0.176} & \textbf{0.270} \\
    \bottomrule
  \end{tabular}
	}
 \caption{Ablation of two-stage MSA's order on three real-world datasets with MSE and MAE metrics.}
	\label{tab:order}
\end{table}

\paragraph{Adapter}
Then, we investigate the impact of dual adapters on model performance. Table \ref{tab:adapter} presents the results of experiments conducted with the removal of adapters from the model architecture. The empirical evidence suggests that the omission of either adapter leads to suboptimal performance outcomes. This phenomenon can be attributed to the inherent dissimilarity like sequence and variable information processed by the model. The interplay between these two types of information is particularly crucial in handling datasets with a higher volume of variables, such as the Electricity dataset. The absence of an adapter markedly deteriorates the predictive accuracy in such scenarios. These findings underscore the critical role of adapters in balancing and effectively integrating diverse data characteristics for enhanced model performance.

\begin{table}[h]
	\centering
	\resizebox{1\linewidth}{!}{
		\begin{tabular}{c|cc|cc|cc}
    \toprule
    \multirow{1}{*}{{Datasets}} & 
    \multicolumn{2}{c}{\rotatebox{0}{{ETTh1}}} &
    \multicolumn{2}{c}{\rotatebox{0}{{Weather}}} &
    \multicolumn{2}{c}{\rotatebox{0}{{Electricity}}} \\
    \cmidrule(lr){2-3} \cmidrule(lr){4-5}\cmidrule(lr){6-7} 
    Metric & {MSE} & {MAE}  & {MSE} & {MAE}  & {MSE} & {MAE} \\
    \toprule
    {w/o all Adapter} & 0.434 & 0.438 & 0.252 & 0.282 & 0.192 & 0.284\\
    {w/o Channel Adapter} & 0.436 & 0.436 & \textbf{0.250} & \textbf{0.280} & 0.201 & 0.291\\
    {w/o Sequence Adapter} & 0.429 & 0.434 & \textbf{0.250} & 0.281 & 0.192 & 0.284\\
    \cmidrule(lr){1-7}
    {CSformer} & \textbf{0.429} &\textbf{0.432} & \textbf{0.250} & \textbf{0.280} & \textbf{0.176} & \textbf{0.270} \\
    \bottomrule
  \end{tabular}
	}
 \caption{Ablation of adapter on three real-world datasets with MSE and MAE metrics.}

	\label{tab:adapter}
\end{table}

\subsection{Model Analysis}

\paragraph{Correlation Analysis}

We show the results of the correlation visualization in the Weather dataset in Figure ~\ref{fig:score}.
\begin{figure}[htbp]
\begin{center}
\includegraphics[width=1\columnwidth]{attention-score.pdf}

\caption{A case visualization
of score maps by two-stage attention.}
\label{fig:score}
\end{center}
\end{figure}
Initially, the blocks exhibit exploratory, uniform attention distribution. However, as the model deepens, attention becomes more focused on specific channels and sequence dimensions, indicating a growing learning of inter-channel and temporal dependencies. Particularly in the final blocks, heightened attention concentration reveals the model's ability to identify key features and moments. This highlights the cross-dimensional information integration facilitated by shared parameters, crucial for understanding the intrinsic representation mechanisms of deep learning models in multivariate time series analysis.

\paragraph{Patch Embedding vs Dimension-augmented Embedding}
 
Dimension-augmented embedding enhances the dimensional representation of sequence data without compromising original information. This approach not only preserves data integrity but also offers processing flexibility, including the option of channel independence or mixing. Such augmentation, a vital preprocessing step, ensures data compatibility with advanced models while enriching its representation to uncover complex patterns and dependencies. 

\begin{figure}[h]
\begin{center}
\includegraphics[width=0.9\columnwidth]{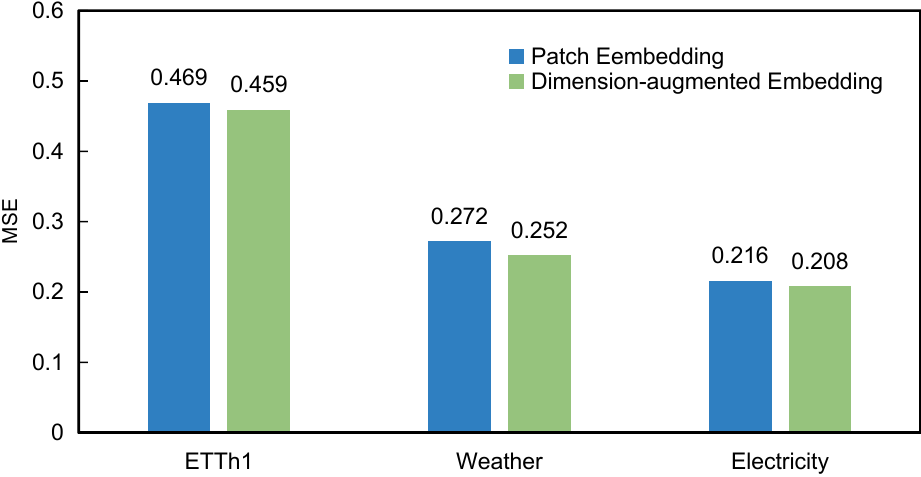}
\caption{Performance of Patch Embedding and Dimension-augmented Embedding.}
\label{fig:patch}
\vspace{-3mm}
\end{center}
\end{figure}

To validate the efficacy of Dimension-augmented Embedding (DE), we replaced the Patch Embedding (PE) in PatchTST ~\citep{PatchTST} with Dimension-augmented Embedding and conducted comparisons across three datasets. Figure ~\ref{fig:patch} compares the performance. It can be seen that our proposed embedding method is generalized and effective.

\paragraph{Generalization}

The Two-stage MSA in CSformer enhances the ability to integrate sequence and channel information from the data. Consequently, the model is capable of learning inherent correlations, which holds promise for improving generalization capability. To investigate this, we trained the model on one dataset and tested it on another. Specifically, we selected datasets with the same sampling frequency, such as ETTh2 to ETTh1. Additionally, we explored datasets with different sampling frequencies, such as ETTh1 to ETTm1.

\begin{table}[htbp]
\centering
\resizebox{1\linewidth}{!}{
\begin{tabular}{c|cccc|cccc}
\toprule
Dataset & \multicolumn{4}{c|}{{ETTh2 \(\to\) ETTh1}} & \multicolumn{4}{c}{{ETTh1 \(\to\) ETTm1}} \\ \cmidrule(lr){2-5} \cmidrule(lr){6-9}
Horizon & {96} & {192} & {336} & {720} & {96} & {192} & {336} & {720} \\ \toprule

{PatchTST} & 0.491 & 0.529 & 0.555 & 0.627 & 0.761 & 0.788 & 0.777 & 0.790 \\
{iTransformer} & 0.880 & 0.920 & 0.924 & 0.919 & 0.923 & 0.938 & 0.938 & 0.945 \\
{CSformer} & \textbf{0.444} & \textbf{0.484} & \textbf{0.512} & \textbf{0.555} & \textbf{0.726} & \textbf{0.776} & \textbf{0.759} & \textbf{0.789} \\
\bottomrule
\end{tabular}
}
\caption{Comparison of generalization capabilities between CSformer and other Transformer-based models. ``Dataset A \(\to\) Dataset B'' indicates training and validation on the training and validation sets of Dataset A, followed by testing on the test set of Dataset B.}

\vspace{-1mm}
\label{tab:domain}
\end{table}

The experimental results, as shown in Table ~\ref{tab:domain}, demonstrate that CSformer outperforms other models in various scenarios. We observed that iTransformer has the poorest generalization performance, which can be attributed to its embedding method in the sequence dimension cannot learn robust representations. When cross-domain data distribution changes, the parameters learned in the source domain become inapplicable. The poor performance of PatchTST can be attributed to its use of a channel-independent method, which fails to integrate hidden correlation between channels and sequences. Therefore, the CSformer model exhibits outstanding generalization capabilities. Additionally, we conducted a supplementary simulation experiment to verify that CSformer is a more robust predictor (see Appendix C).

\paragraph{Training Strategy}
In our earlier discussion, we highlighted the benefits of a training strategy that initially processes channels independently (CI) and then mixes them. This method prevents noise introduced by premature channel mixing (CM) and overcomes the limitations of prolonged independence impeding inter-channel information exchange. To investigate this hypothesis, we conducted an experiment with PatchTST ~\citep{PatchTST}, modifying the dimension of its attention mechanism. Specifically, we shifted the focus of the attention mechanism from the sequence dimension to the channel dimension to facilitate channel mixing. Table ~\ref{tab:civscm} indicates that a reasonable balance of channel independence and mixing enhances predictive performance in multivariate time series forecasting, significantly boosting modeling capabilities. This also provides a new perspective for future research: \textit{how to effectively combine channel independence and mixing?}

\begin{table}[h]
	\centering
	\resizebox{1\linewidth}{!}{
		\begin{tabular}{c|cc|cc|cc}
    \toprule
    \multirow{1}{*}{{Datasets}} & 
    \multicolumn{2}{c}{\rotatebox{0}{{ETTh1}}} &
    \multicolumn{2}{c}{\rotatebox{0}{{Weather}}} &
    \multicolumn{2}{c}{\rotatebox{0}{{Electricity}}} \\
    \cmidrule(lr){2-3} \cmidrule(lr){4-5}\cmidrule(lr){6-7} 
    Metric & {MSE} & {MAE}  & {MSE} & {MAE}  & {MSE} & {MAE} \\
    \toprule
    {CI} & 0.469 & 0.454 & 0.259 & 0.281 & 0.216 & 0.304\\
    {\textbf{CM}} & \textbf{0.442} & \textbf{0.431} & \textbf{0.248} & \textbf{0.278} & \textbf{0.179} & \textbf{0.273} \\
    \cmidrule(lr){1-7}
    {Promotion} & 5.8\% & 5.1\% & 4.2\% & 1.1\% & 17.1\% & 10.2\% \\
    \bottomrule
  \end{tabular}
	}
 \caption{A comprehensive performance comparison between channel independence and channel mixing is presented. The results represent the average values across all four prediction lengths.}
	\label{tab:civscm}
\end{table}
\section{Conclusion and Future Work}
In this article, we summarized the channel-independent and mixing models and show that channel-independent followed by channel mixing is a superior strategy. Based on this, we introduced CSformer, which is an architecture for a two-stage attention mechanism. It balances channel independence and channel mixing to robustly model multivariate time series. In future work, we will investigate more rational ways of combining channel independence and channel mixing and reducing their computational demand.

\appendices
\section{Implementation Details}
\subsection{Dataset Descriptions}
To evaluate the performance of CSformer, we utilized several commonly used datasets across various domains in multivariate long-term forecasting, including ETT (ETTh1, ETTh2, ETTm1, ETTm2), Weather, Electricity, and Solar Energy. These datasets were chosen due to their relevance, comprehensiveness, and the diversity of features they offer.

\textbf{Weather} The Weather dataset comprises 11 weather features and one target variable, "wet bulb." This dataset was collected from nearly 1,600 locations across the United States, spanning 2010 to 2013. The data were sampled at 10-minute intervals, providing a high-resolution temporal dimension. The inclusion of various weather features allows for a robust evaluation of forecasting models in handling complex meteorological data. The "wet bulb" temperature, being a critical measure for understanding human comfort and heat stress, serves as the target variable in our forecasting tasks.

\textbf{Solar Energy} Accurate prediction of solar energy generation is essential for optimizing energy utilization and integrating renewable energy sources into the grid. The Solar Energy dataset contains 137 variables, with a total of 52,179 data points, each sampled at 10-minute intervals. This dataset offers a rich source of information for modeling and predicting solar energy patterns. By leveraging the extensive temporal data, we aim to improve the accuracy and reliability of solar energy forecasts, thereby enhancing the efficiency of energy systems.

\textbf{ETT (Electricity Transformer Temperature)} Monitoring and predicting transformer temperatures are crucial for preventing equipment failures and ensuring the reliability of power systems. The ETT dataset is divided into two categories based on sampling intervals: 15-minute intervals (ETTm1 and ETTm2) and 1-hour intervals (ETTh1 and ETTh2). We used two years of data for our experiments, allowing us to capture both short-term fluctuations and long-term trends in transformer temperatures. This dataset provides a valuable benchmark for assessing the performance of forecasting models in the context of electrical infrastructure monitoring.

\textbf{Electricity} Forecasting electrical consumption load is critical for the effective planning and scheduling of grid maintenance activities. The Electricity dataset used in our study includes 321 variables and three years of historical data collected at hourly intervals. This dataset encompasses a wide range of factors influencing electricity consumption, such as weather conditions, economic activities, and social behaviors. By utilizing this extensive dataset, we aim to develop robust forecasting models that can accurately predict future electricity demand, thereby supporting the efficient management of power grids.

These datasets are widely recognized in the research community for their relevance and robustness in evaluating multivariate forecasting models. By employing these diverse and comprehensive datasets, we aim to thoroughly assess the capabilities of CSformer in handling different types of temporal data and delivering accurate long-term forecasts.
\subsection{Baseline Descriptions}
We have selected a multitude of advanced deep learning models that have achieved SOTA performance in multivariate time series forecasting. Our choice of Transformer-based models includes the Informer ~\citep{Informer}, Autoformer ~\citep{Autoformer}, Stationary ~\citep{Stationary}, FEDformer ~\citep{fedformer}, Crossformer ~\citep{Crossformer}, PatchTST ~\citep{PatchTST}, and the leading model, iTransformer ~\citep{liu2023itransformer}. Notably, Crossformer ~\citep{Crossformer} shares conceptual similarities with our proposed CSformer. Furthermore, recent advancements in MLP-based models have yielded promising predictive outcomes. In this context, we have specifically selected representative models such as DLinear ~\citep{DLinear} and TiDE ~\citep{Tide} for our comparative evaluation. Additionally, for a more comprehensive comparison, we also selected TCN-based models, which include SCINet ~\citep{SCINet} and Timesnet ~\citep{Timesnet}. A brief description of these models is as follows:

\begin{algorithm*}[htbp]
  \caption{CSformer - Overall Architecture.}\label{algo:csformer}
  \begin{algorithmic}[1]
  \Require  
  Input lookback time series $\mathbf{X}\in\mathbb{R}^{N\times L}$; input Length $L$; predicted length $T$; variates number $N$; token dimension $D$; CSformer block number $M$.

    \State $\mathbf{X}=\mathbf{X}.\texttt{unsqueeze}$ \Comment{$\mathbf{X}\in\mathbb{R}^{N\times L\times 1}$}

    \State $\triangleright \ $ Multiply with the learnable vector $ \mathbf{\nu}\in\mathbb{R}^{1\times D}$ to embed $\mathbf{X}$ into the token dimension.
    \State $\mathbf{H}^{0}=\mathbf{X} \times \mathbf{\nu}$ \Comment{$\mathbf{H}^{0}\in\mathbb{R}^{N\times L\times D}$}

    \State $\textbf{for}\ l\ \textbf{in}\ \{1,\cdots,M\}\textbf{:}$\Comment{Run through CSformer blocks.}

    \State $\textbf{\textcolor{white}{for}}$ $\triangleright \ $Channel MSA layer is applied on channel dimension.
    \State $\textbf{\textcolor{white}{for}}\ \mathbf{H}_{c}^{l-1} = \mathbf{H}^{l-1}.\texttt{permute}$
    \Comment{$\mathbf{H}_{c}^{l-1}\in\mathbb{R}^{L\times N\times D}$}
    \State $\textbf{\textcolor{white}{for}}\ \mathbf{Z}_{c}^{l-1} = \texttt{MSA}\big(\mathbf{H}_{c}^{l-1}\big)$
    \Comment{$\mathbf{Z}_{c}^{l-1}\in\mathbb{R}^{L\times N\times D}$}
    \State $\textbf{\textcolor{white}{for}}\ \mathbf{A}_{c}^{l-1} = \texttt{Adapter}\big(\texttt{Norm}(\mathbf{Z}_{c}^{l-1})\big) + \mathbf{H}_{c}^{l-1}$
    \Comment{$\mathbf{A}_{c}^{l-1}\in\mathbb{R}^{L\times N\times D}$}
    
     \State $\textbf{\textcolor{white}{for}}$ $\triangleright \ $Sequence MSA layer is applied on sequence dimension.
    \State $\textbf{\textcolor{white}{for}}\ \mathbf{H}_{s}^{l-1} = \mathbf{A}_{c}^{l-1}.\texttt{permute}$
    \Comment{$\mathbf{H}_{s}^{l-1}\in\mathbb{R}^{N\times L\times D}$}
    \State $\textbf{\textcolor{white}{for}}\ \mathbf{Z}_{s}^{l-1} = \texttt{MSA}\big(\mathbf{H}_{s}^{l-1}\big)$
    \Comment{$\mathbf{Z}_{s}^{l-1}\in\mathbb{R}^{N\times L\times D}$}
    \State $\textbf{\textcolor{white}{for}}\ \mathbf{A}_{s}^{l-1} = \texttt{Adapter}\big(\texttt{Norm}(\mathbf{Z}_{s}^{l-1})\big) + \mathbf{H}_{s}^{l-1}$
    \Comment{$\mathbf{A}_{s}^{l-1}\in\mathbb{R}^{N\times L\times D}$}

    \State $\textbf{\textcolor{white}{for}}$ $\triangleright \ $ $\mathbf{A}_{s}^{l-1}$ as input to the next block $\mathbf{H}^{l}$
    \State $\textbf{\textcolor{white}{for}}\ \mathbf{H}^{l} = \mathbf{A}_{s}^{l-1}$
    \Comment{$\mathbf{H}^{l}\in\mathbb{R}^{N\times L\times D}$}
    \State $\textbf{End for}$

    \State $\mathbf{\hat{X}}=\texttt{Linear}(\mathbf{H}^{M}.\texttt{flatten})$ \Comment{Project to predicted series, $\mathbf{\hat{X}}\in\mathbb{R}^{N\times T}$}
    
    \State $\textbf{Return}\ \mathbf{\hat{X}}$ \Comment{Return the prediction result $\mathbf{\hat{X}}$}
  \end{algorithmic} 
\end{algorithm*}

\begin{itemize}
\item Informer ~\citep{Informer} model is a transformer-based solution for long sequence time-series forecasting with a ProbSparse self-attention mechanism, self-attention distilling, and a generative style decoder for faster inference.

\item Autoformer ~\citep{Autoformer} features a novel decomposition architecture and Auto-Correlation mechanism, moving away from traditional self-attention to efficiently capture long-range dependencies in time series data.

\item Non-stationary Transformers ~\citep{Stationary} employs a dual-module structure for time series forecasting with non-stationary data, including a Series Stationarization module and a De-stationary Attention module to handle varying patterns effectively.

\item FEDformer ~\citep{fedformer} combines Transformers with seasonal-trend decomposition, leveraging sparse frequency representations for linear complexity and improved effectiveness in long-term series forecasting.

\item DLinear ~\citep{DLinear} is a simple yet effective architecture for long-term time series forecasting, challenging the complexity of Transformer-based models with a direct multi-step strategy.

\item TimesNet ~\citep{Timesnet} transforms 1D time series into 2D tensors using TimesBlock for better capturing complex temporal patterns and employing a parameter-efficient inception block for enhanced representation and analysis.

\item TiDE ~\citep{Tide} is an efficient MLP-based encoder-decoder model for long-term time-series forecasting, combining the speed of linear models with the ability to manage non-linear dependencies and covariates.

\item SCINet ~\citep{SCINet} employs a unique recursive architecture, focusing on downsampling and convolution for time series forecasting and interaction with temporal features across multiple resolutions.

\item Crossformer ~\citep{Crossformer} is a Transformer-based model for multivariate time series forecasting, utilizing Dimension-Segment-Wise embedding and a Two-Stage Attention layer within a Hierarchical Encoder-Decoder structure.

\item PatchTST ~\citep{PatchTST} is a Transformer-based model for multivariate time series forecasting with subseries-level patch segmentation and channel independent architecture, optimizing local semantic information retention.

\item iTransformer ~\citep{liu2023itransformer} revises the traditional Transformer for time series forecasting by inverting embedding dimension, focusing on capturing multivariate correlations and learning nonlinear representations for enhanced performance and flexibility.

\end{itemize}

\subsection{Implementation Details}
All experiments were implemented using PyTorch and conducted on a single NVIDIA A40 48GB GPU. For optimization, we employed the ADAM optimizer, setting the initial learning rate to either \(10^{-4}\) or \(1.5 \times 10^{-4}\). The model optimization process utilized the L2 loss function to ensure robust performance. Batch sizes were configured to be either 64 or 128 to balance computational efficiency and model performance.

We varied the number of CSformer blocks \(M\) within the proposed model, selecting values from \(\{1, 2, 3\}\). The dimensionality of the series representation \(D\) was adjusted to \(\{16, 64, 128\}\) to explore the impact of different feature dimensions on model performance. This experimental setup allowed us to thoroughly evaluate the CSformer across a range of configurations and determine the optimal settings for accurate long-term forecasting.

To provide a clear understanding of the CSformer implementation, we included the pseudocode in Algorithm ~\ref{algo:csformer}, which outlines the key steps and operations of the model. This pseudocode serves as a guide for replicating our experiments and understanding the detailed functioning of the CSformer architecture.

\begin{figure*}[htbp]
\includegraphics[width=1\textwidth,height=0.5\textwidth]{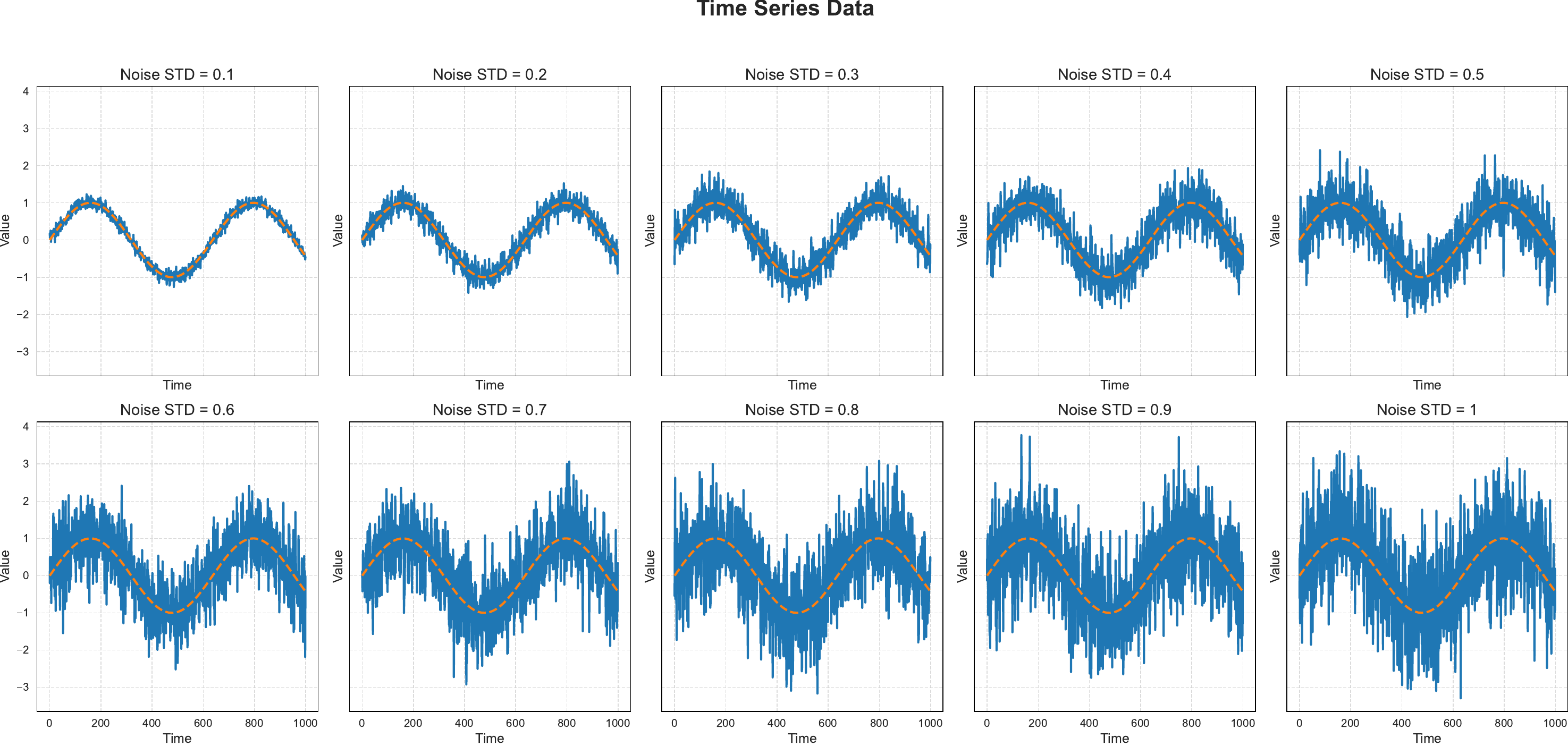}
\caption{Data with different amplitudes (std).}\label{fig:different std}
\end{figure*}

\begin{figure*}[htbp]
\includegraphics[width=1\textwidth,height=0.5\textwidth]{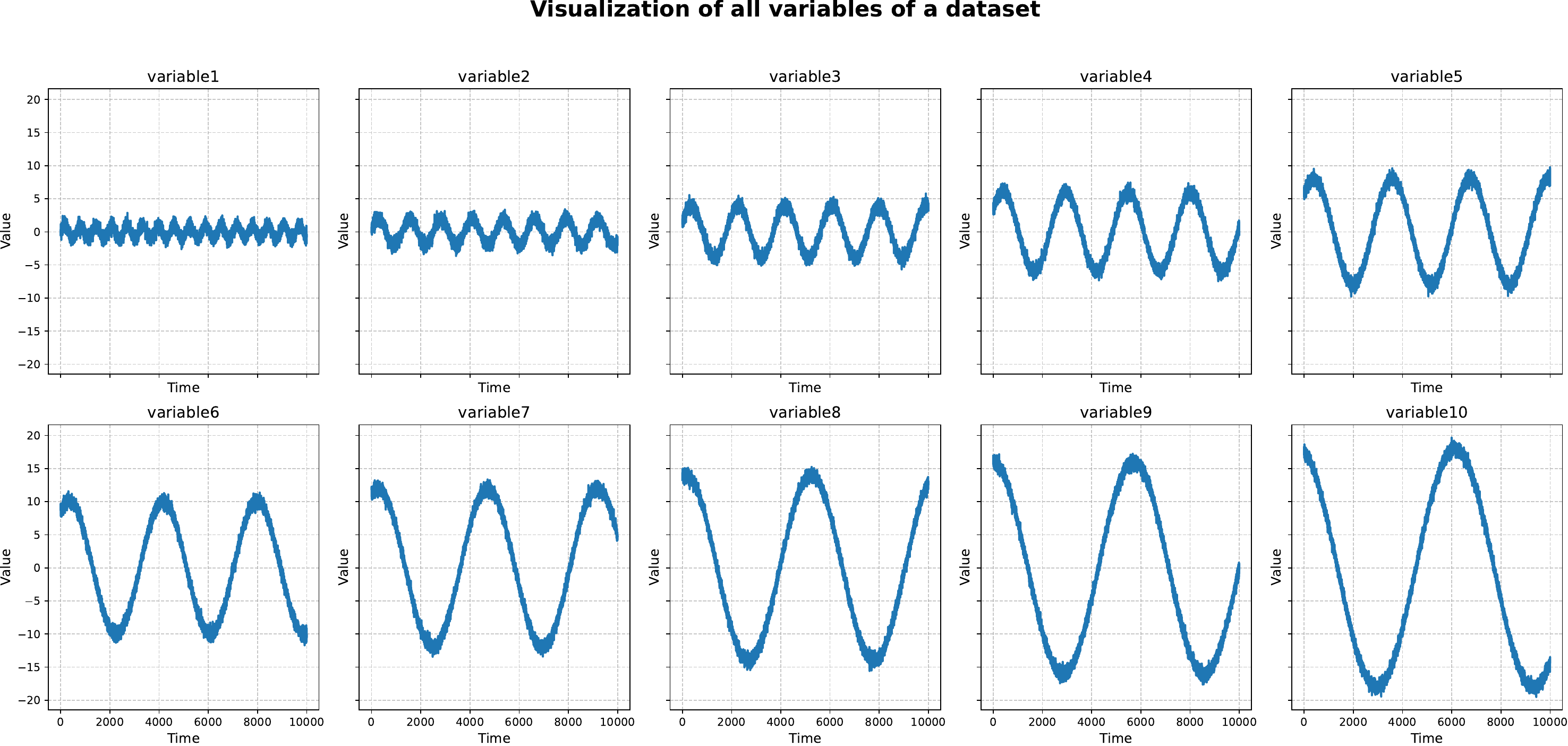}
\caption{Different variables in a dataset.}\label{fig:different variable}
\end{figure*}
\section{Discussions on Embedding Methods}
Currently, the mainstream methods for embedding multivariate time series data can be broadly categorized into three types:

\begin{enumerate}
    \item \textbf{Embedding All Variables at a Single Time Step into One Token:} This is the most common embedding approach, employed by methods such as Informer ~\citep{Informer}, Autoformer ~\citep{Autoformer}, and Fedformer ~\citep{fedformer}. In this method, all variables at a given time step are combined into a single token. However, since the values of these variables represent different physical meanings and are recorded by distinct measurements, embedding them into a single token may obscure the inter-variable correlations. Additionally, due to the nature of time series data, a token formed from a single time step often fails to reveal useful information.
    
    \item \textbf{Embedding All Time Steps of a Single Variable into One Token:} This approach was recently proposed by iTransformer ~\citep{liu2023itransformer}. It addresses some of the issues mentioned above to a certain extent. However, directly embedding the entire sequence of a variable may weaken the temporal information, as time dynamics could be smoothed out. Moreover, this method is primarily suited for predictions within a single dataset and lacks good transferability. Since the embedding is specific to a particular sequence, it cannot learn universal representations. Consequently, when there are changes in data distribution across domains, the parameters learned in the source domain may become unsuitable.
    
    \item \textbf{Segmenting the Sequence into Multiple Sub-sequences and Embedding Each Sub-sequence into One Token:} This method, introduced by PatchTST ~\citep{PatchTST}, involves dividing the sequence into several sub-sequences, thus increasing the dimensionality: \( L \rightarrow \text{seg\_num} \times \text{seg\_len} \). This segmentation facilitates channel independence but may lead to information loss. Rapid trend changes or peak events split across different segments might not be fully captured in the segment-level representation. Additionally, some time series patterns may span multiple time blocks, and if these patterns are disrupted during segmentation, it could lead to fragmented or difficult-to-recognize patterns. Other dimensionality reduction techniques include TimesNet ~\citep{Timesnet}, which uses Fast Fourier Transform (FFT) to identify periodic patterns in the sequence and reshape them into a 2D vector. While FFT emphasizes periodic patterns and low-frequency components, it may overlook high-frequency details. This is useful for capturing long-term trends but may be insufficient for detecting short-term spikes or instantaneous changes.
\end{enumerate}
\section{Robustness Analysis}
To demonstrate that a framework like CSformer is a more robust forecaster, we conducted simulation experiments using synthetic data. These experiments are necessary due to the difficulty of obtaining real, noise-free data. We created several time series datasets for our simulations, each containing 20,000 data points and 10 unique variables. Each variable follows a distinct triangular function with varying amplitudes, phases, and periods. For each dataset, the variables adhere to specific triangular functions, chosen from the sets: amplitude $\{1, 2, 4, 6, 8, 10, 12, 14, 16, 18\}$, phase $\{0, 0.2, 0.4, 0.6, 0.8, 1, 1.2, 1.4, 1.6, 1.8\}$, and period $\{1, 2, 3, 4, 5, 6, 7, 8, 9, 10\}$. Each variable follows its designated trigonometric function, ensuring diversity and complexity in the dataset.

\begin{equation}
\begin{aligned}
Variable_1(t) &= \text{amplitude}_1 \cdot \sin\left(\frac{2\pi \cdot t}{\text{period}_1} + \text{phase}1\right) \\
Variable_2(t) &= \text{amplitude}2 \cdot \sin\left(\frac{2\pi \cdot t}{\text{period}2} + \text{phase}2\right) \\
&\vdots \\
Variable{10}(t) &= \text{amplitude}{10} \cdot \sin\left(\frac{2\pi \cdot t}{\text{period}{10}} + \text{phase}{10}\right)
\end{aligned}
\end{equation}

To emulate real-world conditions, we added Gaussian noise to the datasets. Different standard deviations of Gaussian noise were introduced to the first 90\% of the data points in each dataset, while the remaining 10\% of data points remained noise-free. The datasets were split into training and testing subsets, with the first 90\% used for training and the last 10\% for testing.

Gaussian noise with a mean of zero and varying standard deviations was added to the first 90\% of data in ten datasets, with $\text{std} \in \{0.1, 0.2, 0.3, 0.4, 0.5, 0.6, 0.7, 0.8, 0.9, 1\}$. The resulting dataset curves are shown in Appendix Figure~\ref{fig:different std}. A visual representation of all variables in one of the datasets is presented in Appendix Figure~\ref{fig:different variable}.

\begin{figure}[htbp]
\begin{center}
\includegraphics[width=0.6\columnwidth]{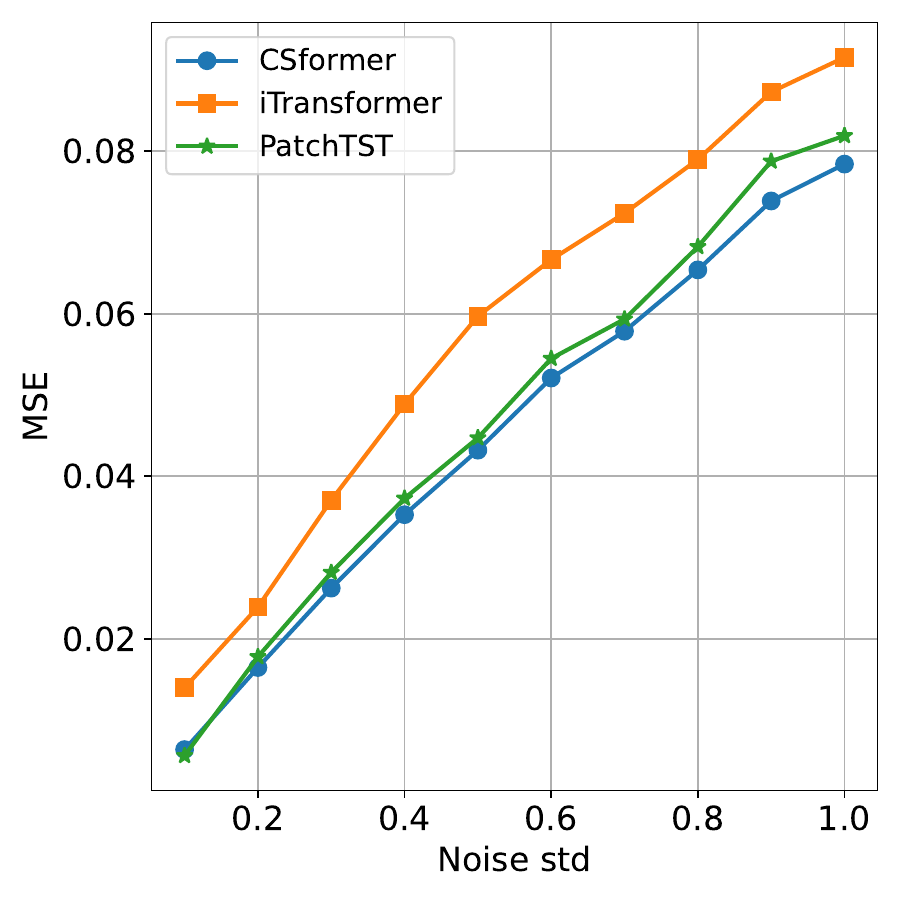}
\caption{The simulation experiment with different amplitudes of Gaussian noises.}
\label{fig:noise}
\end{center}
\end{figure}

From Appendix Figure ~\ref{fig:noise} we observe that the prediction performance of CSformer always outperforms other former classes in each experiment with different noise amplitudes. This illustrates that CSformer is able to extract more implicit correlations from multivariate time series data and can be used as a more robust predictor.

\section{Full Results}

\subsection{Ablation Study}
Additionally, we show the full experimental results of the ablation study.
\begin{table}[h]
	\centering
 \caption{Ablation of two-stage MSA on three real-world datasets with MSE and MAE metrics.}
 \renewcommand{\arraystretch}{1.15}
	\resizebox{1.0\linewidth}{!}{
		\begin{tabular}{cc|c|ccccccccc}
			\toprule
			&\multicolumn{2}{c}{Methods}& \multicolumn{2}{c}{w/o Channel MSA}& \multicolumn{2}{c}{w/o Sequence MSA}& \multicolumn{2}{c}{CSformer}\\
			\cmidrule(r){4-5}\cmidrule(r){6-7}\cmidrule(r){8-9}\cmidrule(r){10-11}
			&\multicolumn{2}{c|}{Metric}&MSE&MAE&MSE&\multicolumn{1}{c}{MAE}&MSE&MAE\\
			\midrule
			&\multirow{4}*{{ETTh1}}& 96    & 0.374 & \textbf{0.397} & 0.385 & \textbf{0.397} & \textbf{0.372} & \textbf{0.397}  \\
            &\multicolumn{1}{c|}{}& 192   & 0.421 & \textbf{0.424} & 0.437 & \textbf{0.424} & \textbf{0.420} & \textbf{0.424}  \\
            &\multicolumn{1}{c|}{}& 336   & 0.458 & 0.442 & 0.476 & 0.442 & \textbf{0.453} & \textbf{0.440}  \\
            &\multicolumn{1}{c|}{}& 720   & \textbf{0.470} & \textbf{0.469} & 0.473 & 0.463 & \textbf{0.470} & 0.470  \\
            \midrule
			&\multirow{4}*{{Weather}}& 96    & \textbf{0.168} & 0.217 & 0.181 & 0.168 & \textbf{0.168} & \textbf{0.217} \\
            &\multicolumn{1}{c|}{}& 192   & \textbf{0.213} & 0.258 & 0.225 & 0.266 & \textbf{0.213} & \textbf{0.257} \\
            &\multicolumn{1}{c|}{}& 336   & \textbf{0.269} & \textbf{0.296} & 0.280 & 0.303 & 0.272 & 0.298 \\
            &\multicolumn{1}{c|}{}& 720   & \textbf{0.343} & \textbf{0.343} & 0.351 & 0.350 & 0.346 & 0.347 \\
			\midrule
			&\multirow{4}*{{Electricity}}& 96    & 0.177 & 0.264 & 0.178 & 0.266 & \textbf{0.146} & \textbf{0.242} \\
            &\multicolumn{1}{c|}{}& 192   & 0.179 & \textbf{0.265} & 0.183 & 0.270 & \textbf{0.172} & 0.266 \\
            &\multicolumn{1}{c|}{}& 336   & 0.196 & 0.282 & 0.198 & 0.286 & \textbf{0.176} & \textbf{0.271} \\
            &\multicolumn{1}{c|}{}& 720   & 0.236 & 0.316 & 0.239 & 0.321 & \textbf{0.211} & \textbf{0.303} \\
            \bottomrule
		\end{tabular}
	}

	\label{tab:MSA}
\end{table}

\begin{table}[htbp]
	\centering
 \caption{Ablation of parameter sharing on three real-world datasets with MSE and MAE metrics.}
 \renewcommand{\arraystretch}{1}
	\resizebox{0.9\linewidth}{!}{
		\begin{tabular}{cc|c|ccccccc}
			\toprule
			&\multicolumn{2}{c}{Methods}& \multicolumn{2}{c}{w/o Parameter Sharing}& \multicolumn{2}{c}{Parameter Sharing}\\
			\cmidrule(r){4-5}\cmidrule(r){6-7}\cmidrule(r){8-9}
			&\multicolumn{2}{c|}{Metric}&MSE&MAE&MSE&\multicolumn{1}{c}{MAE}\\
			\midrule
			&\multirow{4}*{{ETTh1}}& 96    &  0.375 & 0.400 & \textbf{0.372} & \textbf{0.397}  \\
            &\multicolumn{1}{c|}{}& 192   &  0.426 & 0.427 & \textbf{0.420} & \textbf{0.424}  \\
            &\multicolumn{1}{c|}{}& 336   &  0.465 & 0.445 & 
            \textbf{0.453} & \textbf{0.440}  \\
            &\multicolumn{1}{c|}{}& 720   &  0.492 & 0.482 & \textbf{0.470} & \textbf{0.470}  \\
            \midrule
			&\multirow{4}*{{Weather}}& 96   & 0.174 & 0.218 & \textbf{0.168} & \textbf{0.217} \\
            &\multicolumn{1}{c|}{}& 192   & 0.220 & 0.260 & \textbf{0.213} & \textbf{0.257} \\
            &\multicolumn{1}{c|}{}& 336   & 0.277 & 0.299 & \textbf{0.272} & \textbf{0.298} \\
            &\multicolumn{1}{c|}{}& 720  & 0.351 & 0.352 & \textbf{0.346} & \textbf{0.347} \\
			\midrule
			&\multirow{4}*{{Electricity}}& 96   & 0.163 & 0.256 & \textbf{0.146} & \textbf{0.242} \\
            &\multicolumn{1}{c|}{}& 192  & 0.176 & 0.268 & \textbf{0.172} & \textbf{0.266} \\
            &\multicolumn{1}{c|}{}& 336  & 0.189 & 0.282 & \textbf{0.176} & \textbf{0.271} \\
            &\multicolumn{1}{c|}{}& 720   & 0.228 & 0.316 & \textbf{0.211} & \textbf{0.303} \\
            \bottomrule
		\end{tabular}
	}
	\label{tab:paremeter}
\end{table}

\begin{table}[htbp]
	\centering
 \caption{Ablation of two-stage MSA's order on three real-world datasets with MSE and MAE metrics.}
 \renewcommand{\arraystretch}{1}
	\resizebox{0.85\linewidth}{!}{
		\begin{tabular}{cc|c|ccccccc}
			\toprule
			&\multicolumn{2}{c}{Methods}& \multicolumn{2}{c}{S $\longrightarrow $ C}& \multicolumn{2}{c}{C $\longrightarrow $ S}\\
			\cmidrule(r){4-5}\cmidrule(r){6-7}\cmidrule(r){8-9}
			&\multicolumn{2}{c|}{Metric}&MSE&MAE&MSE&\multicolumn{1}{c}{MAE}\\
			\midrule
			&\multirow{4}*{{ETTh1}}& 96    &  0.376 & 0.399 & \textbf{0.372} & \textbf{0.397}  \\
            &\multicolumn{1}{c|}{}& 192   &  0.426 & 0.427 & \textbf{0.420} & \textbf{0.424}  \\
            &\multicolumn{1}{c|}{}& 336   &  0.460 & 0.445 & 
            \textbf{0.453} & \textbf{0.440}  \\
            &\multicolumn{1}{c|}{}& 720   &  0.511 & 0.497 & \textbf{0.470} & \textbf{0.470}  \\
            \midrule
			&\multirow{4}*{{Weather}}& 96   & \textbf{0.168} & 0.218 & \textbf{0.168} & \textbf{0.217} \\
            &\multicolumn{1}{c|}{}& 192   & 0.216 & 0.259 & \textbf{0.213} & \textbf{0.257} \\
            &\multicolumn{1}{c|}{}& 336   & \textbf{0.269} & \textbf{0.296} & 0.272 & 0.298 \\
            &\multicolumn{1}{c|}{}& 720  & \textbf{0.345} & \textbf{0.347} & 0.346 & \textbf{0.347} \\
			\midrule
			&\multirow{4}*{{Electricity}}& 96   & 0.148 & 0.246 & \textbf{0.146} & \textbf{0.242} \\
            &\multicolumn{1}{c|}{}& 192  & \textbf{0.164} & \textbf{0.260} & 0.172 & 0.266 \\
            &\multicolumn{1}{c|}{}& 336  & 0.188 & 0.284 & \textbf{0.176} & \textbf{0.271} \\
            &\multicolumn{1}{c|}{}& 720   & 0.213 & 0.305 & \textbf{0.211} & \textbf{0.303} \\
            \bottomrule
		\end{tabular}
	}
	\label{tab:order}
\end{table}

\paragraph{Two-stage MSA}
Here is a complete ablation study for the two-stage MSA, which includes: (i) removing any single MSA (including the Adapter); (ii) whether parameters are shared or not; (iii) the order of sequential MSA and channel MSA.

\begin{table}[htbp]
	\centering
 \caption{Ablation of adapter on three real-world datasets with MSE and MAE metrics.}
 \renewcommand{\arraystretch}{1.15}
	\resizebox{1.0\linewidth}{!}{
		\begin{tabular}{cc|c|cccccccc}
			\toprule
			&\multicolumn{2}{c}{Methods}& \multicolumn{2}{c}{w/o all Adapter}& \multicolumn{2}{c}{w/o Channel Adapter}& \multicolumn{2}{c}{w/o Sequence Adapter}& \multicolumn{2}{c}{CSformer}\\
			\cmidrule(r){4-5}\cmidrule(r){6-7}\cmidrule(r){8-9}\cmidrule(r){10-11}
			&\multicolumn{2}{c|}{Metric}&MSE&MAE&MSE&MAE&MSE&MAE&MSE&MAE\\
			\midrule
			&\multirow{4}*{{ETTh1}}& 96    & 0.379 & 0.400    & 0.377 &0.399 & 0.374 & 0.398 & \textbf{0.372} & \textbf{0.397}  \\
            &\multicolumn{1}{c|}{}& 192 & 0.425    & 0.427  & 0.425 & 0.427 & 0.422 & 0.425 & \textbf{0.420} & \textbf{0.424}  \\
            &\multicolumn{1}{c|}{}& 336 & 0.462    & 0.447  & 0.468 & 0.447 & 0.459 & 0.444 & \textbf{0.453} & \textbf{0.440}  \\
            &\multicolumn{1}{c|}{}& 720 & 0.472    & 0.476  & 0.474 & 0.473 & \textbf{0.462} & \textbf{0.468} & 0.470 & 0.470  \\
            \midrule
			&\multirow{4}*{{Weather}}& 96 & \textbf{0.168}    & 0.218   & \textbf{0.168} & 0.218 & 0.170 & 0.219 & \textbf{0.168} & \textbf{0.217} \\
            &\multicolumn{1}{c|}{}& 192  & 0.220    & 0.262 & 0.216 & 0.258 & 0.215 & 0.258 & \textbf{0.213} & \textbf{0.257} \\
            &\multicolumn{1}{c|}{}& 336 & 0.275   & 0.302  & 0.273 & 0.299 & \textbf{0.271} & 0.299 & 0.272 & \textbf{0.298} \\
            &\multicolumn{1}{c|}{}& 720 & 0.345    & 0.347  & \textbf{0.343} & \textbf{0.344} & 0.344 & 0.347 & 0.346 & 0.347 \\
			\midrule
			&\multirow{4}*{{Electricity}}& 96 & 0.162    & 0.255   & 0.172 & 0.265 & 0.162 & 0.255 & \textbf{0.146} & \textbf{0.242} \\
            &\multicolumn{1}{c|}{}& 192 & 0.175    & 0.267  & 0.184 & 0.276 & 0.175 & 0.267 & \textbf{0.172} & \textbf{0.266} \\
            &\multicolumn{1}{c|}{}& 336 & 0.198    & 0.291  & 0.194 & 0.286 & 0.198 & 0.291 & \textbf{0.176} & \textbf{0.271} \\
            &\multicolumn{1}{c|}{}& 720 & 0.232    & 0.321  & 0.254 & 0.337 & 0.232 & 0.321 & \textbf{0.211} & \textbf{0.303} \\
            \bottomrule
		\end{tabular}
	}
	\label{tab:adapter}
\end{table}

\paragraph{Adapter}
We also conducted ablation experiments on the Adapter and found that it improves prediction accuracy while minimally increasing the number of parameters.

\begin{table*}[htbp]
  
  \centering
    \renewcommand{\arraystretch}{1.2}
	\resizebox{\linewidth}{!}{
  \begin{tabular}{c|c|cc|cc|cc|cc|cc|cc|cc|cc|cc|cc|cc|cc}
    \toprule
    
    \multicolumn{2}{c}{\multirow{2}{*}{Models}} & 
    \multicolumn{2}{c}{\rotatebox{0}{{\textbf{CSformer}}}} &
    \multicolumn{2}{c}{\rotatebox{0}{{iTransformer}}} &
    \multicolumn{2}{c}{\rotatebox{0}{{PatchTST}}} &
    \multicolumn{2}{c}{\rotatebox{0}{{Crossformer}}} &
    \multicolumn{2}{c}{\rotatebox{0}{{SCINet}}} &
    \multicolumn{2}{c}{\rotatebox{0}{{TiDE}}} &
    \multicolumn{2}{c}{\rotatebox{0}{{{TimesNet}}}} &
    \multicolumn{2}{c}{\rotatebox{0}{{DLinear}}} &
    \multicolumn{2}{c}{\rotatebox{0}{{FEDformer}}} &
    \multicolumn{2}{c}{\rotatebox{0}{{Stationary}}} &
    \multicolumn{2}{c}{\rotatebox{0}{{Autoformer}}} &
    \multicolumn{2}{c}{\rotatebox{0}{{Informer}}} \\
    
    \cmidrule(lr){3-4} \cmidrule(lr){5-6}\cmidrule(lr){7-8} \cmidrule(lr){9-10}\cmidrule(lr){11-12}\cmidrule(lr){13-14} \cmidrule(lr){15-16} \cmidrule(lr){17-18} \cmidrule(lr){19-20} \cmidrule(lr){21-22} \cmidrule(lr){23-24} \cmidrule(lr){25-26}
    \multicolumn{2}{c}{Metric}  & {MSE} & {MAE}  & {MSE} & {MAE}  & {MSE} & {MAE}  & {MSE} & {MAE}  & {MSE} & {MAE}  & {MSE} & {MAE} & {MSE} & {MAE} & {MSE} & {MAE} & {MSE} & {MAE} & {MSE} & {MAE} & {MSE} & {MAE} & {MSE} & {MAE}\\
    \toprule
    \multirow{5}{*}{\update{\rotatebox{90}{\scalebox{0.95}{ETTm1}}}}
    &  {96} & \textbf{{0.324}} & \textbf{{0.367}} &\underline{{0.334}} & \underline{{0.368}} & {0.329} & \textbf{{0.367}} & {0.404} & {0.426} & {0.364} & {0.387} &{{0.338}} &{{0.375}} &{{0.345}} &{{0.372}} & {0.418} & {0.438} &{0.379} &{0.419} &{0.386} &{0.398} &{0.505} &{0.475} &{0.672} &{0.571} \\
    & {192} & \underline{{0.369}} & {0.388} & {0.377} & {0.391}  & \textbf{{0.367}} & \textbf{{0.385}} & {0.450} & {0.451} &{0.398} & {0.404} &{0.374} &\underline{{0.387}}  &{{0.380}} &{{0.389}} & {0.439} & {0.450}  &{0.426} &{0.441} &{0.459} &{0.444} &{0.553} &{0.496} &{0.795} &{0.669}\\
    & {336} & \textbf{{0.396}} & \textbf{{0.408}} & {0.426} & {0.420}  & \underline{{0.399}} & \underline{{0.410}} & {0.532}  &{0.515} & {0.428} & {0.425} &{{0.410}} &{0.411}  &{{0.413}} &{{0.413}} & {0.490} & {0.485}  &{0.445} &{0.459} &{0.495} &{0.464} &{0.621} &{0.537} &{1.212} &{0.871} \\
    & {720} & \textbf{{0.451}} & \textbf{{0.439}} & {0.491} & {0.459}  & \underline{{0.454}} & \textbf{{0.439}} & {0.666} & {0.589} & {0.487} & {0.461} &{{0.478}} &\underline{{0.450}} &{0.474} &{{0.453}} & {0.595} & {0.550}  &{0.543} &{0.490} &{0.585} &{0.516} &{0.671} &{0.561} &{1.166} &{0.823} \\
    \cmidrule(lr){2-26}
    & {Avg} & \textbf{{0.385}} & \textbf{{0.400}} & {0.407} & {0.410}  & \underline{{0.387}} & \textbf{{0.400}} & {0.513} & {0.496} & {0.419} & {0.419} &{0.400} &\underline{{0.406}} &{{0.403}} &{{0.407}} & {0.485} & {0.481}  &{0.448} &{0.452} &{0.481} &{0.456} &{0.588} &{0.517} &{0.961} &{0.734} \\
    \midrule
    \multirow{5}{*}{\update{\rotatebox{90}{\scalebox{0.95}{ETTm2}}}}
    &  {96} & \underline{{0.179}} & {0.269} & {0.180} & \underline{{0.264}}  & \textbf{{0.175}} & \textbf{{0.259}} & {0.287} & {0.366} & {0.207} & {0.305} &{{0.187}} &{0.267} &{0.193} &{0.292} & {0.286} & {0.377} &{0.203} &{0.287} &{{0.192}} &{0.274} &{0.255} &{0.339}  &{0.365} &{0.453} \\
    & {192} & \underline{{0.244}} & \underline{{0.309}} & {0.250} & \underline{{0.309}} & \textbf{{0.241}} & \textbf{{0.302}} & {0.414} & {0.492} & {0.290} & {0.364} &{{0.249}} &{{0.309}} &{0.284} &{0.362} & {0.399} & {0.445} &{0.269} &{0.328} &{0.280} &{0.339} &{0.281} &{0.340} &{0.533} &{0.563} \\
    & {336} & \textbf{{0.303}} & \underline{{0.346}}& {{0.311}} & {{0.348}}  & \underline{{0.305}} & \textbf{{0.343}}  & {0.597} & {0.542}  & {0.377} & {0.422} &{{0.321}} &{{0.351}} &{0.369} &{0.427} & {0.637} & {0.591} &{0.325} &{0.366} &{0.334} &{0.361} &{0.339} &{0.372} &{1.363} &{0.887} \\
    & {720} & \textbf{{0.400}} & \textbf{{0.400}}& {0.412} & {0.407}  & \underline{{0.402}} & \textbf{{0.400}} & {1.730} & {1.042} & {0.558} & {0.524} &{{0.408}} &\underline{{{0.403}}} &{0.554} &{0.522} & {0.960} & {0.735} &{0.421} &{0.415} &{0.417} &{0.413} &{0.433} &{0.432} &{3.379} &{1.338} \\
    \cmidrule(lr){2-26}
    & {Avg} & \underline{{0.282}} &\underline{{0.331}}& {{0.288}} & {{0.332}}  & \textbf{{0.281}} & \textbf{{0.326}} & {0.757} & {0.610} & {0.358} & {0.404} &{{0.291}} &{{0.333}} &{0.350} &{0.401} & {0.571} & {0.537} &{0.305} &{0.349} &{0.306} &{0.347} &{0.327} &{0.371} &{1.410} &{0.810} \\
    \midrule
    \multirow{5}{*}{\rotatebox{90}{\update{\scalebox{0.95}{ETTh1}}}}
    &  {96} & \textbf{{{0.372}}} & \textbf{{{0.394}}}& {{0.386}} & {{0.405}}  & {0.414} & {0.419} & {0.423} & {0.448} & {0.479}& {0.464}  &{0.384} &{{0.402}} & {0.386} &\underline{{0.400}} & {0.654} & {0.599} &\underline{{0.376}} &{0.419} &{0.513} &{0.491} &{0.449} &{0.459}  &{0.865} &{0.713} \\
    & {192} & \textbf{{{0.420}}} & \textbf{{{0.425}}}& {0.441} & {0.436} & {0.460} & {0.445} & {0.471} & {0.474}  & {0.525} & {0.492} &\underline{{0.436}} &\underline{{0.429}}  &{{0.437}} &{{0.432}} & {0.719} & {0.631} &\textbf{{0.420}} &{0.448} &{0.534} &{0.504} &{0.500} &{0.482} &{1.008} &{0.792} \\
    & {336} & \textbf{{{0.453}}} & \textbf{{{0.440}}}& {{0.487}} & \underline{{0.458}} & {0.501} & {0.466} & {0.570} & {0.546} & {0.565} & {0.515} &{0.491} &{0.469} &{{0.481}} & {{0.459}} & {0.778} & {0.659} &\underline{{0.459}} &{{0.465}} &{0.588} &{0.535} &{0.521} &{0.496} &{1.107} &{0.809} \\
    & {720} & \textbf{{{0.470}}} &\textbf{{{0.470}}}& {{0.503}} & {{0.491}} &\underline{{0.500}} & \underline{{0.488}} & {0.653} & {0.621} & {0.594} & {0.558} &{0.521} &{{0.500}} &{0.519} &{0.516} & {0.836} & {0.699} &{{0.506}} &{{0.507}} &{0.643} &{0.616} &{{0.514}} &{0.512}  &{1.181} &{0.865} \\
    \cmidrule(lr){2-26}
    & {Avg} & \textbf{{{0.429}}} & \textbf{{{0.432}}}& {{0.454}} & \underline{{0.447}} & {0.469} & {0.454} & {0.529} & {0.522} & {0.541} & {0.507} &{0.458} &{{0.450}} &{{0.456}} &{{0.452}} & {0.747} & {0.647} &\underline{{0.440}} &{0.460} &{0.570} &{0.537} &{0.496} &{0.487}  &{1.040} &{0.795} \\
    \midrule
    \multirow{5}{*}{\rotatebox{90}{\scalebox{0.95}{ETTh2}}}  
    &  {96} &{\textbf{0.293}} & {\textbf{0.340}} &\underline{{0.297}} & {0.349} & {0.302} & \underline{{0.348}} & {0.745} & {0.584} & {0.707} & {0.621} &{0.400} & {0.440}  & {{0.340}} & {{0.374}} &{{0.333}} &{{0.387}}  &{0.358} &{0.397} &{0.476} &{0.458} &{0.346} &{0.388} &{3.755} &{1.525} \\
    & {192} & {\textbf{0.375}} & {\textbf{0.390}} & \underline{{0.380}} & \underline{{0.400}} &{0.388} & \underline{{0.400}} & {0.877} & {0.656} & {0.860} & {0.689} & {0.528} & {0.509} & {{0.402}} & {{0.414}} &{0.477} &{0.476} &{{0.429}} &{{0.439}} &{0.512} &{0.493} &{0.456} &{0.452} &{5.602} &{1.931} \\
    & {336} & {\textbf{0.378}} & {\textbf{0.406}} & {0.428} & \underline{{0.432}} & \underline{{0.426}} & {0.433}& {1.043} & {0.731} & {1.000} &{0.744} & {0.643} & {0.571}  & {{0.452}} & {{0.452}} &{0.594} &{0.541} &{0.496} &{0.487} &{0.552} &{0.551} &{{0.482}} &{0.486} &{4.721} &{1.835} \\
    & {720} & {\textbf{0.409}} & {\textbf{0.432}}&  \underline{{0.427}} & \underline{{0.445}} & {0.431} & {0.446} & {1.104} & {0.763} & {1.249} & {0.838} & {0.874} & {0.679} & {{0.462}} & {{0.468}} &{0.831} &{0.657} &{{0.463}} &{{0.474}} &{0.562} &{0.560} &{0.515} &{0.511} &{3.647} &{1.625} \\
    \cmidrule(lr){2-26}
    & {Avg} & {\textbf{0.364}} & {\textbf{0.392}}
    & \underline{{0.383}} & \underline{{0.407}} & {0.387} & \underline{{0.407}} & {0.942} & {0.684} & {0.954} & {0.723} & {0.611} & {0.550}  &{{0.414}} &{{0.427}} &{0.559} &{0.515} &{{0.437}} &{{0.449}} &{0.526} &{0.516} &{0.450} &{0.459} &{4.431} &{1.729} \\
    \midrule
    \multirow{5}{*}{\rotatebox{90}{\scalebox{0.95}{Electricity}}} 
    & {96} & {\textbf{0.146}} & \underline{{0.242}} & \underline{{0.148}} & {\textbf{0.240}} & {0.195} & {0.285} & {0.219} & {0.314} & {0.247} & {0.345} & {0.237} & {0.329} &{{0.168}} &{{0.272}} &{0.197} &{0.282} &{0.193} &{0.308} &{{0.169}} &{{0.273}} &{0.201} &{0.317}  &{0.274} &{0.368} \\
    & {192} & \underline{{0.172}} & \underline{{0.266}} & {\textbf{0.162}} & {\textbf{0.253}} & {0.199} & {0.289} & {0.231} & {0.322} & {0.257} & {0.355} & {0.236} & {0.330} &{{0.184}} &{0.289} &{0.196} &{{0.285}} &{0.201}  &{0.315} &{{0.182}} &{{0.286}} &{0.222} &{0.334} &{0.296} &{0.386} \\
    & {336} & {\textbf{0.176}} & \underline{{0.271}} & \underline{{0.178}} & {\textbf{0.269}} & {0.215} & {0.305} & {0.246} & {0.337} & {0.269} & {0.369} & {0.249} & {0.344} &{{0.198}} &{{0.300}} &{0.209} &{{0.301}} &{0.214} &{0.329} &{{0.200}} &{0.304} &{0.231} &{0.338}  &{0.300} &{0.394} \\
    & {720} & {\textbf{0.211}} & {\textbf{0.303}} & {0.225} & \underline{{0.317}} & {0.256} & {0.337} & {0.280} & {0.363} & {0.299} & {0.390} & {0.284} & {0.373} &\underline{{{0.220}}} &{{0.320}} &{0.245} &{0.333} &{0.246} &{0.355} &{{0.222}} &{{0.321}} &{0.254} &{0.361} &{0.373} &{0.439} \\
    \cmidrule(lr){2-26}
    & {Avg} & {\textbf{0.176}} & {\textbf{0.270}} & \underline{{0.178}} & \textbf{{0.270}} & {0.216} & {0.304} & {0.244} & {0.334} & {0.268} & {0.365} & {0.251} & {0.344} &{{0.192}} &\underline{{{0.295}}} &{0.212} &{0.300} &{0.214} &{0.327} &{{0.193}} &{{0.296}} &{0.227} &{0.338} &{0.311} &{0.397} \\
    \midrule
    \multirow{5}{*}{\rotatebox{90}{\scalebox{0.95}{Solar Energy}}} 
    &  {96} &{\textbf{0.197}} &\underline{{0.244}} &\underline{{0.203}} &\textbf{{0.237}} & {0.234} & {0.286} &{0.310} &{0.331} &{0.237} &{0.344} &{0.312} &{0.399} &{0.250} &{0.292} &{0.290} &{0.378} &{0.242} &{0.342} &{0.215} &{0.249} &{0.884} &{0.711} &{0.236} &{0.259} \\
    & {192} &\underline{{0.230}} &{0.274} &{0.233} &\textbf{{0.261}} & {0.267} & {0.310} &{0.734} &{0.725} &{0.280} &{0.380} &{0.339} &{0.416} &{0.296} &{0.318} &{0.320} &{0.398} &{0.285} &{0.380} &{0.254} &{0.272} &{0.834} &{0.692} &\textbf{{0.217}} &\underline{{0.269}} \\
    & {336} &{\textbf{0.244}} &\underline{{0.281}} &\underline{{0.248}} &\textbf{{0.273}} & {0.290}  &{0.315} &{0.750} &{0.735} &{0.304} &{0.389} &{0.368} &{0.430} &{0.319} &{0.330} &{0.353} &{0.415} &{0.282} &{0.376} &{0.290} &{0.296} &{0.941} &{0.723} &{0.249} &{0.283}\\
    & {720} &\underline{{0.247}} &\underline{{0.280}} &{0.249} &\textbf{{0.275}} &{0.289} &{0.317} &{0.769} &{0.765} &{0.308} &{0.388} &{0.370} &{0.425} &{0.338} &{0.337} &{0.356} &{0.413} &{0.357} &{0.427} &{0.285} &{0.295} &{0.882} &{0.717} &\textbf{{0.241}} &{0.317}\\
    \cmidrule(lr){2-26}
    & {Avg} &{\textbf{0.230}} &\underline{{0.270}}&\underline{{0.233}} &{\textbf{0.262}} &{0.270} &{0.307} &{0.641} &{0.639} &{0.282} &{0.375} &{0.347} &{0.417} &{0.301} &{0.319} &{0.330} &{0.401} &{0.291} &{0.381} &{0.261} &{0.381} &{0.885} &{0.711} &{0.235} &{0.280}\\
    \midrule
    \multirow{5}{*}{\rotatebox{90}{\scalebox{0.95}{Weather}}} 
    &  {96} & \underline{{0.168}} & \underline{{0.217}} & {0.174} & {\textbf{0.214}} & {0.177} & {0.218} & \textbf{{0.158}} & {0.230} & {0.221} & {0.306} & {0.202} & {0.261} &{{0.172}} &{{0.220}} & {0.196} &{0.255} & {0.217} &{0.296} & {{0.173}} &{{0.223}} & {0.266} &{0.336} & {0.300} &{0.384}  \\
    & {192} & {\underline{0.213}} & {\underline{0.257}} & {0.221} & {\textbf{0.254}} & {0.225} & {0.259} & {\textbf{0.206}} & {0.277} & {0.261} & {0.340} & {0.242} & {0.298} &{{0.219}} &{{0.261}}  & {0.237} &{0.296} & {0.276} &{0.336} & {0.245} &{0.285} & {0.307} &{0.367} & {0.598} &{0.544} \\
    & {336} & {\textbf{0.272}} & {0.298} & {\underline{0.278}} & {\textbf{0.296}} & {\underline{0.278}} & {\underline{0.297}} & {\textbf{0.272}} & {0.335} & {0.309} & {0.378} & {0.287} & {0.335} &{{0.280}} &{{0.306}} & {0.283} &{0.335} & {0.339} &{0.380} & {0.321} &{0.338} & {0.359} &{0.395} &{0.578} &{0.523} \\
    & {720} & {\textbf{0.346}} & {\textbf{0.347}} & {0.358} & {0.349} & {\underline{0.354}} & {\underline{0.348}} & {0.398} & {0.418} & {0.377} & {0.427} & {0.351} & {0.386} &{0.365} &{{0.359}} & {{0.345}} &{{0.381}} & {0.403} &{0.428} & {0.414} &{0.410} & {0.419} &{0.428} & {1.059} &{0.741} \\
    \cmidrule(lr){2-26}
    & {Avg} & {\textbf{0.250}} & {\underline{0.280}} & {\underline{0.258}} & {\textbf{0.279}} & {0.259} & {0.281} & {0.259} & {0.315} & {0.292} & {0.363} & {0.271} & {0.320} &{{0.259}} &{{0.287}} &{0.265} &{0.317} &{0.309} &{0.360} &{0.288} &{0.314} &{0.338} &{0.382} &{0.634} &{0.548} \\
    \midrule
     \multicolumn{2}{c|}{{{$1^{\text{st}}$ Count}}} & {\textbf{26}} & {\textbf{18}} & {1} & {\underline{12}} & {\underline{4}} & {9} & {3} & {0} & {0} & {0} & {0} & {0} & {0} & {0} & {0} & {0} & {1} & {0} & {0} & {0} & {0} & {0} & {2} & {0}\\
    \bottomrule
  \end{tabular}
  }\caption{Full results for the multivariate forecasting task. We compare extensive competitive models under different prediction lengths following the setting of iTransformer. The input sequence length is set to 96 for all baselines. \emph{Avg} means the average results from all four prediction lengths. The best results are in \textbf{bold} and the second best are \underline{underlined}.}\label{tab:full_baseline_results}
\end{table*}

\subsection{Main Results}
Due to the limitations on the length of the main text, the complete results of the multivariate time series forecasting will be presented in this section. Appendix Table ~\ref{tab:full_baseline_results} shows the complete results, with CSformer achieving the best predictions among the many models.

\clearpage

\bibliography{aaai25}

\end{document}